\documentclass{article}

\usepackage{microtype}
\usepackage{graphicx}
\usepackage{subfigure}
\usepackage{booktabs}

\usepackage[title]{appendix}

\usepackage{times}
\usepackage{epsfig}
\usepackage{graphicx}
\usepackage{float}
\usepackage{wrapfig}
\usepackage{amsmath,amssymb,amsthm}
\usepackage{bm,xspace}
\usepackage{comment}
\usepackage{verbatim}
\usepackage{multirow}
\usepackage{balance}
\usepackage{url}
\usepackage{booktabs}
\usepackage{etoolbox,siunitx}
\usepackage{calc}
\usepackage{pifont,hologo}
\newcommand{\ra}[1]{\renewcommand{\arraystretch}{#1}}

\usepackage[super]{nth}
\usepackage{nicefrac}
\robustify\bfseries

\def\dd{\mathbf{d}}
\def\ee{\mathbf{e}}
\def\ff{\mathbf{f}}

\def\hh{\mathbf{h}}

\def\pp{\mathbf{p}}
\def\qq{\mathbf{q}}
\def\rr{\mathbf{r}}

\def\tt{\mathbf{t}}

\def\vv{\mathbf{v}}

\def\xx{\mathbf{x}}

\def\zz{\mathbf{z}}

\def\EE{\mathbf{E}}

\def\GG{\mathbf{G}}

\def\II{\mathbf{I}}

\def\LL{\mathbf{L}}
\def\MM{\mathbf{M}}

\def\QQ{\mathbf{Q}}
\def\RR{\mathbf{R}}

\def\TT{\mathbf{T}}

\DeclareMathOperator*{\minimize}{minimize}

\def\trans{^{\top}}

\DeclareMathSymbol{@}{\mathord}{letters}{"3B}

\newcommand\mypara[1]{\vspace{1mm}\noindent\textbf{#1}}

\def\latex/{\LaTeX}
\def\bibtex/{\hologo{BibTeX}}

\usepackage{amsmath,amsthm,amsfonts,amssymb,amscd}
\usepackage{enumerate}
\usepackage{fancyhdr}
\usepackage{mathrsfs}
\usepackage{xcolor}
\usepackage{graphicx}
\usepackage{listings}
\usepackage{amsmath}
\usepackage{hyperref}
\usepackage{mathtools}
\usepackage{balance}

\newcommand{\PD}[2]{\frac{\partial{#1}}{\partial{#2}}}

\usepackage{hyperref}

\usepackage[accepted]{icml2020}

\icmltitlerunning{Scalable Differentiable Physics for Learning and Control}

\begin{document}

\twocolumn[
\icmltitle{Scalable Differentiable Physics for Learning and Control}

\begin{icmlauthorlist}
\icmlauthor{Yi-Ling Qiao}{umd}
\icmlauthor{Junbang Liang}{umd}
\icmlauthor{Vladlen Koltun}{intel}
\icmlauthor{Ming C. Lin}{umd}
\end{icmlauthorlist}

\icmlaffiliation{umd}{University of Maryland, College Park}
\icmlaffiliation{intel}{Intel Labs}

\icmlcorrespondingauthor{Yi-Ling Qiao, Junbang Liang, and Ming C. Lin}{\{yilingq,liangjb,lin\}@cs.umd.edu}
\icmlcorrespondingauthor{Vladlen Koltun}{vladlen.koltun@intel.com}

\icmlkeywords{Differentiable Simulation}

\vskip 0.3in
]
\printAffiliationsAndNotice{}
\begin{abstract}
Differentiable physics is a powerful approach to learning and control problems that involve physical objects and environments. While notable progress has been made, the capabilities of differentiable physics  solvers remain limited. We develop a scalable framework for differentiable physics that can support a large number of objects and their interactions. To accommodate objects with arbitrary geometry and topology, we adopt meshes as our representation and leverage the sparsity of contacts for scalable differentiable collision handling. Collisions are resolved in localized regions to minimize the number of optimization variables even when the number of simulated objects is high. We further accelerate implicit differentiation of optimization with nonlinear constraints. Experiments demonstrate that the presented framework requires up to two orders of magnitude less memory and computation in comparison to recent particle-based methods. We further validate the approach on inverse problems and control scenarios, where it outperforms derivative-free and model-free baselines by at least an order of magnitude. 
\end{abstract}

\section{Introduction}
\label{sec:introduction}
Differentiable physics enables gradient-based learning systems to strictly adhere to physical dynamics. By making physics simulation differentiable, we can backpropagate through the physical consequences of actions. This enables agents to quickly learn to achieve desired effects in the physical world. It is also a principled and effective approach to inverse problems that involve physical systems~\citep{Degrave2019,Belbute2018,Toussaint2018differentiable,Schenck2018}.

Recent efforts have significantly advanced the understanding of differentiable physics in machine learning and robotics. Automatic differentiation and analytical methods have been applied to derive a variety of differentiable simulation engines~\cite{ingraham2019learning,Liang2019,Holl2020,Hu2019:ICLR}. Yet existing differentiable frameworks are still limited in their ability to simulate complex scenes composed of many detailed interacting objects. For example, modeling assumptions have limited some existing frameworks to restricted object classes, such as balls or two-dimensional polygons~\citep{Belbute2018,Degrave2019}. Other approaches adopt expressive grid- or particle-based formulations, which are general in principle but have not scaled to detailed simulation of large scenes due to the cubic growth rates of volumetric grids and particles~\cite{Mrowca2018,Hu2019:ICRA,Li2019}. For example, a grid of size $1024^3$ has on the order of a billion elements, and this resolution is still quite coarse in comparison to the scale and detail of physical environments we encounter in daily life, such as city streets. Particle-based representations likewise suffer from explosive growth rates or limited resolution: \citet{Macklin2014} suggest a maximum size ratio of 1:10 between the smallest and largest feature, or placing particles only around boundaries, which runs the risk of object tunneling.

In this paper, we develop a differentiable physics framework that addresses these limitations. We adopt \emph{meshes} as a general representation of objects. Meshes are the most widely used specification for object geometry in computer graphics and scientific computing. Meshes are inherently sparse, can model objects of any shape, and can compactly specify environments with both large spatial extent and highly detailed features~\cite{Botsch2010}.

The use of meshes brings up the challenge of collision handling, since collisions can occur anywhere on the surface of the mesh. Prior frameworks adopted global LCP-based formulations, which severely limited scalability~\citep{Degrave2019,Belbute2018}. In contrast, we leverage the structure of contacts by grouping them into localized impact zones, which can be efficiently represented and processed~\cite{Bridson2002,Harmon2008}. This substantially reduces the number of variables and constraints involved in specifying dynamical scenes and dramatically speeds up backpropagation. 

Following prior work, we use implicit differentiation to compute gradients induced by an optimization problem embedded in the simulation~\cite{AmosKolter2017,Liang2019}. One of our contributions is an acceleration scheme that can handle the nonlinear constraints encountered in our system.

We evaluate the scalability and generality of the presented approach experimentally. Controlled experiments that vary the number of objects in the scene and their relative scales demonstrate that our approach is dramatically more efficient than state-of-the-art differentiable physics frameworks. Since meshes can represent both rigid and deformable objects, our work is also the first differentiable physics framework that can simulate two-way coupling of rigid bodies and cloth. We further demonstrate example applications to learning and control scenarios, where the presented framework outperforms derivative-free and model-free baselines by at least an order of magnitude. Code is available on our project page: \url{https://gamma.umd.edu/researchdirections/mlphysics/diffsim}

\section{Related Work}
\label{sec:related}

Backpropagating through physical dynamics is an enticing prospect for embedding physical reasoning into learning processes.
\citet{Degrave2019} articulated the potential of this approach and applied automatic differentiation to rigid body systems where contacts are assumed to only happen between balls and planes. Although this assumption simplifies collision detection and response, it greatly limits the scope of application. \citet{Belbute2018} introduced implicit differentiation for gradient computation, obviating the need to explicitly differentiate through all steps of the forward optimization solver. By designing specific rules for collision response, their framework supports simple two-dimensional shapes such as circles and polygons, but was not extended to general 3D objects. \citet{Toussaint2018differentiable} utilized differentiable primitives to develop a robot reasoning system that can achieve user-defined tasks. It is designed for path planning with sphere-swept shapes and frictionless interactions, but not for general control tasks.

Building on the implicit differentiation methodology, \citet{Liang2019} developed a differentiable simulator for cloth. Their formulation assumes that each node can move independently and is only tied to others by soft constraints. This is true for cloth but not for rigid bodies.
\citet{CarpentierMansard2018} developed algorithms for efficient computation of analytical derivatives of rigid-body dynamics, but did not handle collision. \citet{Millard2020} implemented an articulated body simulator in a reverse-mode automatic differentiation framework, but likewise did not handle dynamic contact and collision response, which are at the heart of our work.

\citet{Hu2019:ICRA} developed a differentiable simulator for deformable objects based on the material point method (MPM). A follow-up work applies this framework to controller design for soft robots~\cite{Spielberg2019}. This approach is effective in handling soft bodies, but has limited ability to enforce rigid structure. Since the method is based on particles and grids, it has limited scalability and has not been applied to large scenes with interacting objects. In contrast, our method uses meshes to represent objects, naturally enforces precise detailed geometry, and scales to scenes with both large spatial extent and fine local features.

\citet{Hu2019:ICLR} introduce a domain-specific language for building differentiable physical simulators. This work develops a programming language and compiler that can be broadly useful throughout the differentiable physics community, including to projects such as ours. However, at present this framework does not deal with general mesh-based rigid and deformable objects. Their rigid-body scenario uses rectangles constrained by springs, with contact responses only possible between rectangles and the ground plane. Their three-dimensional scenario uses soft bodies modeled by particles and grids and inherits the aforementioned limitations of MPM-based modeling.

A line of work developed particle-based frameworks for differentiable simulation of fluids~\cite{Schenck2018,Ummenhofer2020} and unified treatment of fluids and solids~\cite{Mrowca2018,Li2019}. These frameworks resort to approximations in handling rigid bodies and detailed geometry, and suffer from high growth rates in the scale of objects and scenes. The generality of particle-based approaches is appealing, but their scalability limitations are well-known and are explicitly discussed in the literature~\cite{Macklin2014}. We take a different tack and model objects as meshes, thus harnessing the scalability and generality of this representation.

Some of the aforementioned works, as well as others~\cite{battaglia2016interaction,chang2017compositional,Sanchez-Gonzalez2018,Li2019:ICRA}, fit function approximators to demonstrations of physical dynamics. Such approximators are valuable, but may not fully capture the true underlying physics and may degenerate outside the training distribution. Our work follows the program exemplified by \citet{Degrave2019}, \citet{Belbute2018}, and others in that we aim to backpropagate through the true physical dynamics. In this approach, physical correctness is enforced by construction.

\section{Overview}
\label{sec:overview}
\begin{figure*}
    \centering
    \includegraphics[width=1\linewidth]{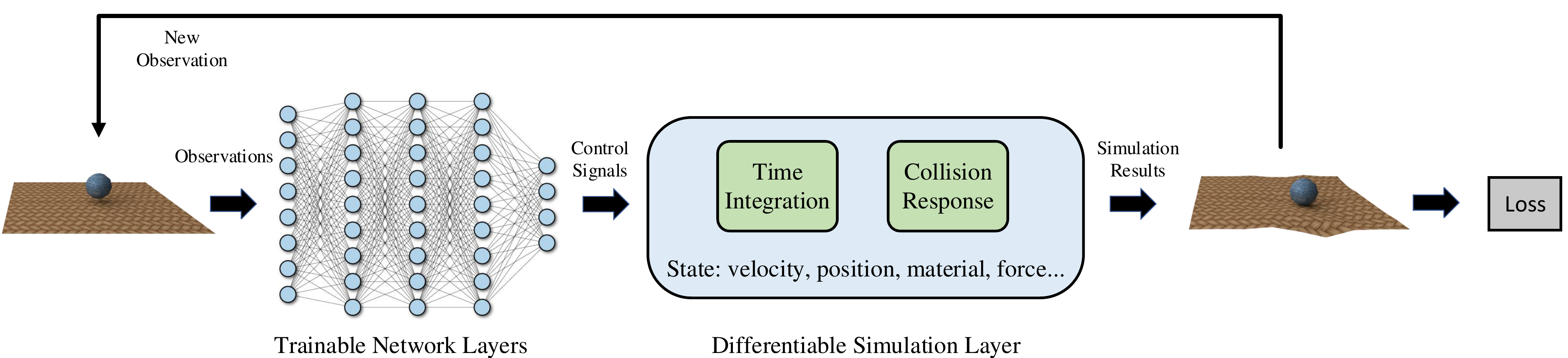}
    \vspace{-4mm}
    \caption{Differentiable simulation embedded in a neural network. The loss can be backpropagated through the physics simulator to the control signal, thus enabling end-to-end gradient-based training of the controller network. This approach enables faster convergence with higher accuracy in learning and control of physical systems.}
    \label{fig:flowchart}
\vspace{-3mm}
\end{figure*}

Our differentiable physics engine can function as a layer embedded within a broader differentiable pipeline, with other neural network layers in the loop. This scenario is illustrated in Figure~\ref{fig:flowchart}. Given observations from simulation, the network can compute control signals and send them to the simulator. The physics engine performs time integration, evaluates collision response, and updates the mesh states. The simulation results can be used to evaluate the loss and provide new observations to the network, thus closing the loop.
All operations in our physics engine are differentiable, supporting end-to-end gradient-based optimization of the complete pipeline.
As mentioned earlier, state-of-the-art differentiable simulation algorithms often suffer from scalability issues. They either use particle representations, which have cubic growth rates with respect to object sizes, or compute custom dynamic responses for a limited number of object primitives, which have limited ability to represent complex shapes. Our method resolves these issues to advance scalability and generality.

Section~\ref{sec3-1} introduces our notation and the basic forward integration scheme.
To facilitate scalability and generality, we adopt meshes as our basic object representation.
Simulation of complex scenes brings up the challenge of collision detection and response. We describe an efficient collision handling scheme in Section~\ref{sec3-2}. Observing that collisions are often sparse, we utilize the impact zone method and handle collisions and contact forces locally in independent collision areas. 
This collision handling scheme has linear computational complexity with respect to the number of constraints rather than the total number of nodes.
Our unified collision handling formulation also enables differentiable two-way coupling of rigid bodies and cloth.
In Section~\ref{sec3-3}, we describe the challenges involved in differentiation and introduce an acceleration scheme for implicit differentiation with nonlinear constraints.

\section{Preliminaries}
\label{sec3-1}

During the presentation of our method, the configuration of the whole system is collected into one vector ${\bm{\qq}=[\bm{\qq}_1\trans,\bm{\qq}_2\trans,...,\bm{\qq}_n\trans]\trans}$, where $\bm{\qq}_{1\leq i \leq n}$ are the generalized coordinates of all objects in the system. Similarly, we use $\bm{\dot{\qq}}=[\bm{\dot{\qq}}_1\trans,\bm{\dot{\qq}}_2\trans,...,\bm{\dot{\qq}}_n\trans]\trans$ to denote the velocity of the corresponding coordinates. For a rigid body, the generalized coordinates $\bm{\qq}_i$ have 6 degrees of freedom (DOF), corresponding to position and orientation. Accordingly, $\bm{\dot{\qq}}_i$ can be interpreted as a vector of linear and angular velocities. For a cloth node, $\bm{\qq}_j$ has 3 DOF, corresponding to its Cartesian coordinates, and $\bm{\dot{\qq}}_j$ is the velocity vector.

After construction of the system, the dynamics can be written as
\begin{equation}
\frac{d}{dt}
\left(
\begin{array}{c}
\bm{\qq}\\
\bm{\dot{\qq}}
\end{array}\right)=
\left(
\begin{array}{c}
\bm{\dot{\qq}}\\
\bm{\ddot{\qq}}
\end{array}\right)=
\left(
\begin{array}{c}
\bm{\dot{\qq}}\\
\bm{\MM}^{-1}\bm{f}(\bm{\qq},\bm{\dot{\qq}})
\end{array}\right),
\end{equation}
where $\bm{\MM}$ is the mass matrix and $\bm{f}$ is the generalized force vector. We incorporate contact forces~\cite{Harmon2008}, internal forces including bending and stretching~\cite{Narain2012}, and external forces including gravity and control input. For numerical simulation, the dynamical system can be discretized with time step $h$. At time $t_0$, with $\bm{\qq}_0=\bm{\qq}(t_0)$ and $\bm{\dot{\qq}}_0=\bm{\dot{\qq}}(t_0)$, we can compute the increment of the coordinates $\Delta\bm{\qq}=\bm{\qq}(t_0+h)-\bm{\qq}(t_0)$ and velocities $\Delta\bm{\dot{\qq}}=\bm{\dot{\qq}}(t_0+h)-\bm{\dot{\qq}}(t_0)$ using the implicit Euler method~\cite{Witkin1997}:
\begin{equation}
\left(
\begin{array}{c}
\Delta\bm{\qq}\\
\Delta\bm{\dot{\qq}}
\end{array}\right)=
h
\left(
\begin{array}{c}
\bm{\dot{\qq}}_0+\Delta\bm{\dot{\qq}}\\
\bm{M}^{-1}\bm{f}(\bm{\qq}+\Delta\bm{\qq},\bm{\dot{\qq}}+\Delta\bm{\dot{\qq}})
\end{array}\right).
\end{equation}
With a linear approximation, we can solve for $\Delta\bm{\dot{q}}$,
\begin{equation}
\left(h^{-1}\bm{\MM}-\PD{\bm{f}}{\bm{\dot{\qq}}}-h\PD{\bm{f}}{\bm{\qq}}\right)\Delta\bm{\dot{\qq}}=\bm{f}_0+h\PD{\bm{f}}{\bm{\qq}}\bm{\dot{\qq}}_0,
\label{eq:implict}
\end{equation}
which can then be used to compute $\bm{\qq}(t_0+h)$ and $\bm{\dot{\qq}}(t_0+h)$ for the next time step.

\section{Scalable Collision Handling}
\label{sec3-2}

One of the most time-consuming and complex parts of physics simulation is collision handling. The difficulty, especially for differentiable physics, lies in how to compute the correct derivatives efficiently even in a large scene with many interacting objects. As observed by \citet{Hu2019:ICLR}, naive discrete-time impulse-based collision response can lead to completely incorrect gradients. We apply continuous collision detection to circumvent this problem. We also employ a bounding volume hierarchy to localize and accelerate dynamic collision detection.

Traditional collision response algorithms for rigid bodies commonly employ global LCP-based solvers to resolve collisions~\cite{Belbute2018}. This works well for small-to-medium-size systems, but does not scale well when the number of objects increases. The problem is exacerbated when the derivatives of all DOFs in this large system need to be computed simultaneously, which is exactly what backpropagation does.

\begin{figure}
\centering
\includegraphics[width=1\linewidth]{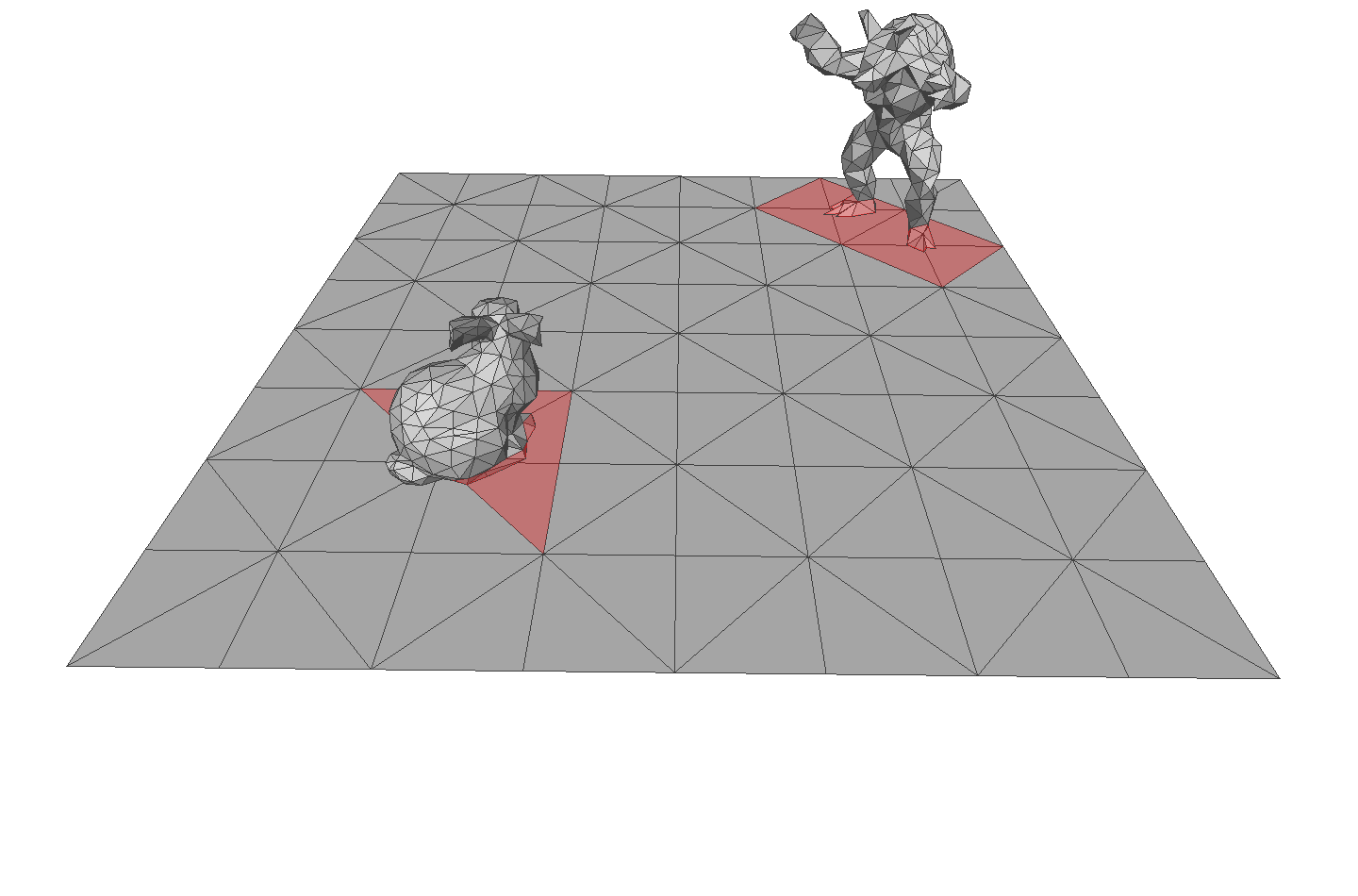}
\vspace{-21mm}
\caption{Collision visualization. Regions in collision are shown in red. In most cases, the number of collisions is sparse compared to the number of degrees of freedom in the scene.}
\vspace{-3mm}
\label{fig:sparsecollision}
\end{figure}

Our key observation is that collisions are sparse in the majority of simulation steps, which means that the number of collisions is much smaller than the total DOF. As shown in Figure~\ref{fig:sparsecollision}, collisions are commonly localized around the scene. With this observation, we resolve collision locally among the objects involved, rather than setting up a global optimization, updating coordinates, and computing the gradients for all objects in the environment.

\begin{wrapfigure}[6]{r}[0pt]{0.45\linewidth}
\centering
\vspace{-5mm}
\includegraphics[width=1\linewidth]{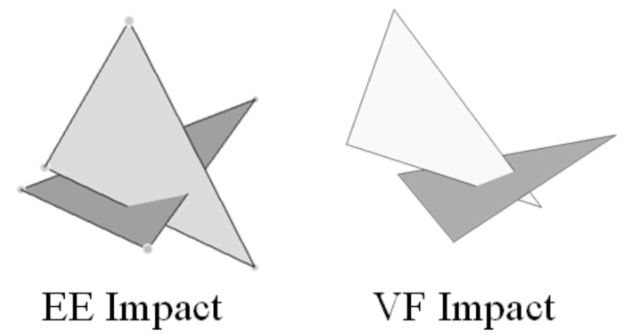}
\label{fig:impact}
\end{wrapfigure}

One way to represent local collisions is to use impact zones~\cite{Bridson2002,Harmon2008}. An impact is a pair of primitives colliding with each other. It can be an edge-edge (EE) or a vertex-face (VF) pair. For these two cases, the non-penetration constraints can be written as
\begin{align}
    0\leq C_{ee}&=\bm{n}\cdot[(\alpha_3\bm{x}_3+\alpha_4\bm{x}_4)-(\alpha_1\bm{x}_1+\alpha_2\bm{x}_2)]\notag \\
    0\leq C_{vf}&=\bm{n}\cdot[\bm{x}_4-(\alpha_1\bm{x}_1+\alpha_2\bm{x}_2+\alpha_3\bm{x}_3)],
\label{eq:constraint}
\end{align}
respectively, where $\xx_i$ are the positions of the mesh vertices, $\alpha_i$ are the barycentric coordinates, and $\bm{n}$ is the normal of the colliding plane.

Impacts may share vertices. All the impacts in one connected component are said to form an impact zone. Each impact zone is a local area that can be treated independently. The red areas in Figure~\ref{fig:sparsecollision} are impact zones.
Collision resolution in one impact zone can be written as an optimization problem:
\begin{alignat}{2}
& \underset{\xx'}{\minimize}\quad &&
\frac{1}{2}(\xx-\xx')\trans\MM(\xx-\xx') \notag\\
& \text{subject to}\quad &&
\GG\xx'+\hh\leq\bm{0}.
\label{eq:impact}
\end{alignat}
Here $\xx$ is the concatenation of vertex positions in the impact zone before collision handling; $\xx'$ are the resolved positions that satisfy constraints; $\MM$ is the mass matrix; $\GG$ and $\hh$ are a matrix and a vector derived from the constraints in Equation~\ref{eq:constraint}. 

The impact zone method is fail-safe~\cite{Bridson2002} and is commonly used in cloth simulation, where each vertex in the cloth is free and can be explicitly optimized when solving Equation~\ref{eq:impact}. However, we cannot directly use this formulation for rigid bodies, because all vertices in a rigid body are tied and cannot be optimized separately. A possible treatment is to use additional constraints on the relative positions of vertices to enforce rigidity. However, the number of vertices in a rigid body is usually far larger than its DOF. We therefore take a different approach that minimizes the number of optimization variables in the system. This is critical for scalable backpropagation in such systems.

We make a distinction between contact vertices $\xx$ and the objects' generalized coordinates $\qq$. The actual optimization variables are replaced by $\qq$. For a rigid body $i$ with rotation $\rr=(\phi,\theta,\psi)\trans$ and translation $\tt=(t_x,t_y,t_z)\trans$, the generalized coordinates are $\qq=[\rr\trans,\tt\trans]\trans\in\mathbb{R}^6$. The block-diagonal mass matrix $\hat{\MM}\in\mathbb{R}^{6\times 6}$ is accordingly changed into angular and linear inertia. 
The constraints are rewritten using the new variables. A function $\bm{f}(\cdot)$ maps generalized coordinates $\qq$ to a contact point $\xx$, including both rotation and translation. 
Detailed formulations of $\hat{\MM}$ and $\bm{f}(\cdot)$ are given in Appendices~\ref{sup:mass} and~\ref{sup:trans}. The optimization problem for collision resolution becomes
\begin{alignat}{2}
&\underset{\qq'}{\minimize}\quad &&
\frac{1}{2}(\qq-\qq')\trans\hat{\MM}(\qq-\qq') \notag\\
& \text{subject to}\quad &&
\GG\bm{f}(\qq')+\hh\leq\bm{0}.
\label{eq:reform}
\end{alignat}

\section{Fast Differentiation}
\label{sec3-3}

Gradients for the sparse linear system in Equation~\ref{eq:implict} can be computed via implicit differentiation~\cite{AmosKolter2017,Liang2019}. Other basic operations can be handled by automatic differentiation~\cite{Paszke2019}. However, the heaviest computation is induced by collision handling, which can take up to 90\% of the runtime for a single simulation step. Collision resolution can involve multiple iterations of the optimization specified in Equation~\ref{eq:reform} for each impact zone. \citet{Liang2019} proposed a method that reduces computation, but it only deals with linear constraints and does not apply in our case due to the the nonlinearity introduced by $\bm{f}(\cdot)$. We now develop an appropriate acceleration scheme that works in our setting.

Consider the differentiation in our case. At a local minimum $\zz^*$ with Lagrange multiplier $\lambda^*$, the following KKT conditions hold:
\begin{align}
    \hat{\MM}\zz^*-\hat{\MM}\qq+\nabla\bm{f}\trans\GG\trans\lambda^*&=0 \notag\\
    D(\lambda^*)(\GG\bm{f}(\zz^*)+\hh)&=0,
\label{eq:kkt}
\end{align}
where $D(\cdot)$ packages a vector into a diagonal matrix. The key idea here is to linearize $\bm{f}(\cdot)$ around a small neighborhood of $\zz^*$. The implicit differentiation is formulated as
\begin{align}
&\left[
\begin{array}{cc}
\hat{\MM} & \nabla\bm{f}\trans\GG\trans\\
D(\lambda^*)\GG\nabla\bm{f} & D(\GG\bm{f}(\zz^*)+\hh)
\end{array}\right]
\left[
\begin{array}{c}
\partial\zz\\
\partial\lambda
\end{array}\right]
=\notag \\
&\left[
\begin{array}{c}
\hat{\MM}\partial\qq-\nabla\bm{f}\trans\partial\GG\trans\lambda^*\\
-D(\lambda^*)(\partial\GG\cdot\bm{f}(\zz^*)+\partial\hh)
\end{array}\right].
\end{align}

As derived by \citet{AmosKolter2017}, by solving for $\dd_\zz,\dd_\lambda$ in the following equation
\begin{equation}
\left[
\begin{array}{cc}
\hat{\MM} & \nabla\bm{f}\trans\GG\trans D(\lambda^*)\\
\GG\nabla\bm{f} & D(\GG\bm{f}(\zz^*)+\hh)
\end{array}\right]
\left[
\begin{array}{c}
\dd_\zz\\
\dd_\lambda
\end{array}\right]
=
\left[
\begin{array}{c}
\PD{\mathcal{L}}{\zz}\trans\\
\bm{0}
\end{array}\right]
\label{eq:system}
\end{equation}
we can get the backward gradients:
\begin{align}
\PD{\mathcal{L}}{\qq}&=\dd_\zz\trans\hat{\MM}\\
\PD{\mathcal{L}}{\GG}&=-D(\lambda^*)\dd_\lambda\ff(\zz^*)\trans-\lambda^*\dd_\zz\trans\nabla\ff\trans\\
\PD{\mathcal{L}}{\hh}&=-\dd_\lambda\trans D(\lambda^*).
\end{align}

Assume that in one impact zone, there are $n$ DOFs and $m$ constraints. In Equation~\ref{eq:system}, the size of the system is $n+m$. The solution of this system would typically take $O((n+m)^3)$: prohibitively expensive in large scenes. We therefore need to accelerate the computation of $\dd_\zz$ and $\dd_\lambda$. \citet{Liang2019} proposed to use a QR decomposition. However, it only copes with quadratic programming with linear constraints.
We can extend this method to nonlinear constraints by incorporating the Jacobian $\nabla\bm{f}$.

Recall that a rigid body has coordinates $\qq=[\rr,\tt]\in\mathbb{R}^6$. The rotation is represented by Euler angles $\rr=(\phi,\theta,\psi)$.
The Jacobian at vertex $\xx=(x,y,z)\trans$ of the rigid body is
\begin{equation}
 \nabla\bm{f}_k=
\left[
\begin{array}{cccccc}
\PD{x}{\phi}    & \PD{x}{\theta}  & \PD{x}{\psi} & 1 & 0 & 0\\
\PD{y}{\phi} & \PD{y}{\theta}    & \PD{y}{\psi}  & 0 & 1 & 0\\
\PD{z}{\phi}  & \PD{z}{\theta} & \PD{z}{\psi}    & 0 & 0 & 1
\end{array}
\right].
\label{eq:grad}
\end{equation}
The exact formulation of $\nabla\bm{f}_k$ can be found in
Appendix~\ref{sup:gradient}.

After computing $\nabla\bm{f}$, we apply a QR decomposition $\sqrt{\hat{\MM}}^{-1}\nabla\bm{f}\trans\GG\trans=\QQ\RR$ and derive $\dd_\zz,\dd_\lambda$ from Equation~\ref{eq:system} as
\begin{align}
\dd_\zz&=\sqrt{\hat{\MM}}^{-1}(\II-\QQ\QQ\trans)\sqrt{\hat{\MM}}^{-1}\PD{\LL}{\zz}\trans\\
\dd_\lambda&=D(\lambda^*)^{-1}\RR^{-1}\QQ\trans\sqrt{\hat{\MM}}^{-1}\PD{\LL}{\zz}\trans.
\end{align}

The cost of the QR decomposition is $O(nm^2)$, reducing the computation from $O((n+m)^3)$.

\section{Experiments}
\label{sec:experiments}
\begin{figure*}
\centering
\begin{tabular}{@{}c@{\hspace{1mm}}c@{\hspace{1mm}}c@{\hspace{1mm}}c@{\hspace{1mm}}c@{}}
   
    \includegraphics[width=0.31\linewidth]{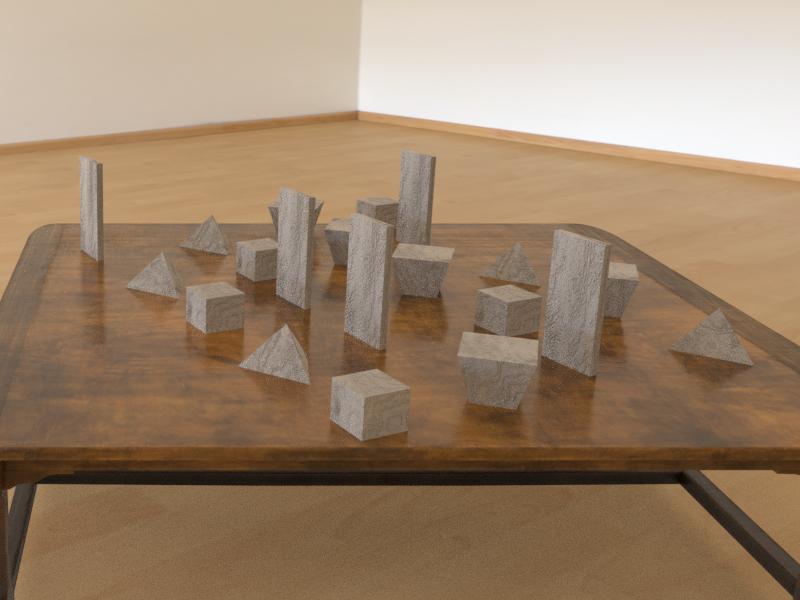} &
    \includegraphics[width=0.31\linewidth]{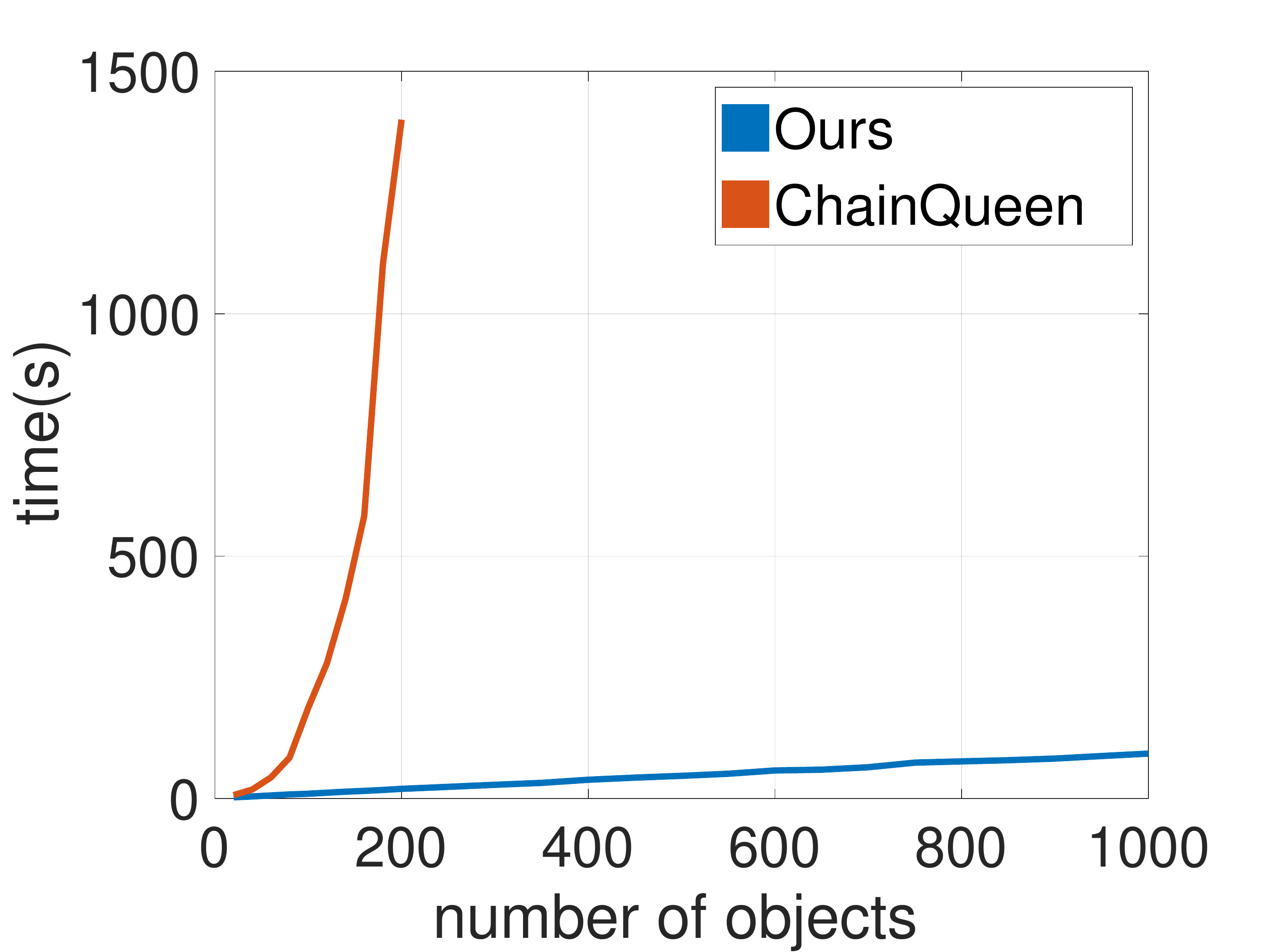} &
    \includegraphics[width=0.31\linewidth]{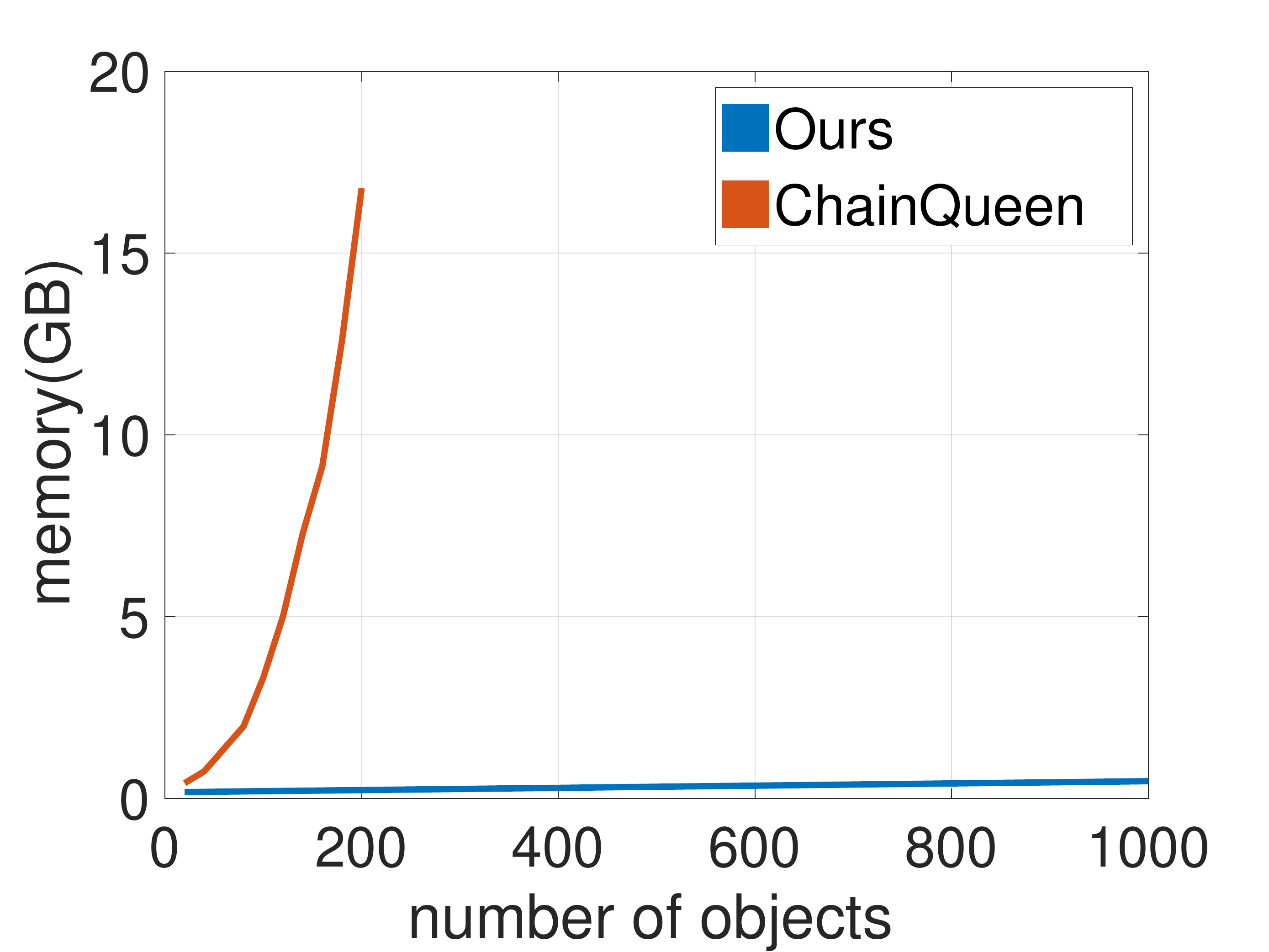} \\
    \includegraphics[width=0.31\linewidth]{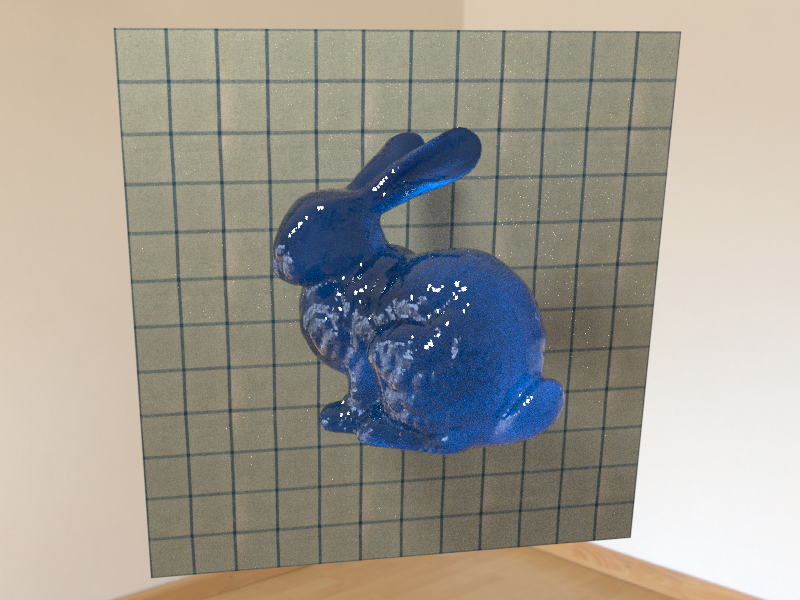} &
    \includegraphics[width=0.31\linewidth]{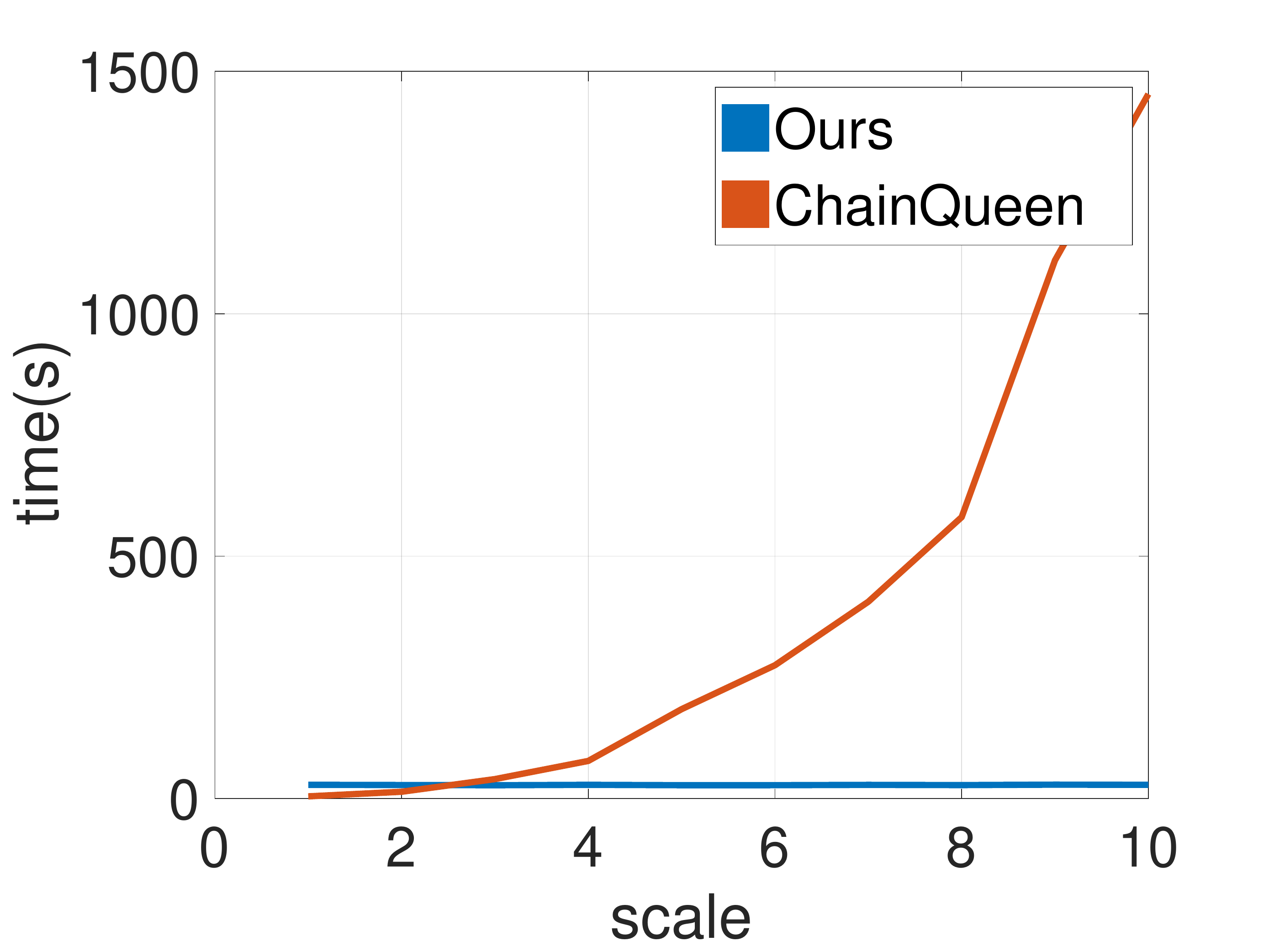} &
    \includegraphics[width=0.3\linewidth]{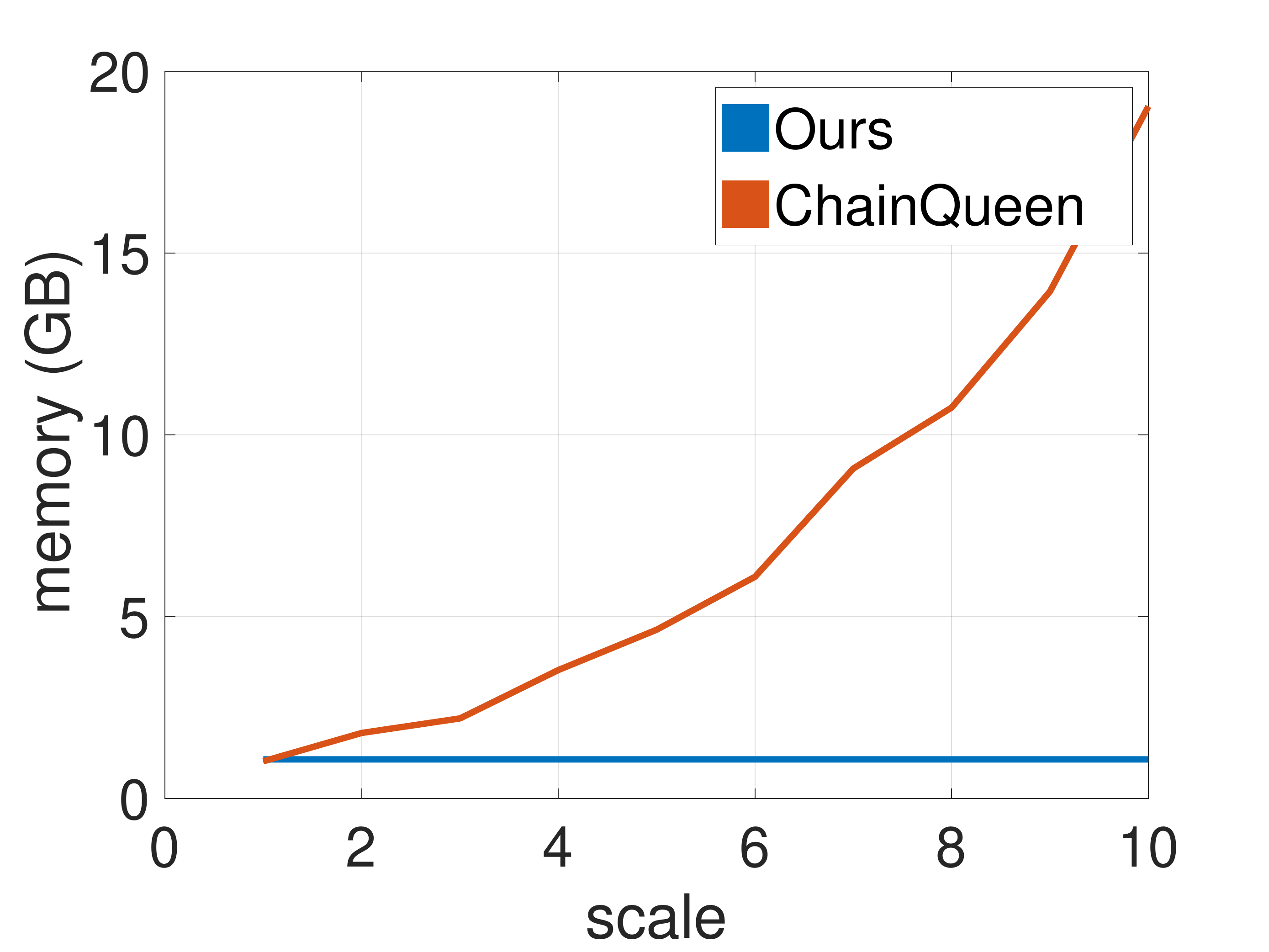} \\
   \small (a) Benchmark scenes   & \small (b) Running time  & \small (c) Memory consumption  
\end{tabular}
\caption{Scalability. (a) Benchmark scenes. Few differentiable simulation frameworks are capable of modeling these scenes. We compare to the expressive MPM-based framework ChainQueen \citep{Hu2019:ICRA}, implemented in the high-performance DiffTaichi library \citep{Hu2019:ICLR}. (b,c) Runtime and memory consumption as the scenes are varied in controlled fashion. Memory consumption is peak memory usage. Time is the running time for simulating 2 seconds of dynamics. Top: the number of objects in the scene increases from 20 to 1000. Bottom: the relative scale of the cloth and the bunny increases from 1:1 to 10:1. The MPM-based framework consumes up to two orders of magnitude more memory and computation before it runs out of memory. 
}
\vspace{-3mm}
\label{fig:scale}
\end{figure*}

We conduct a variety of experiments to evaluate the presented approach. We begin by comparing to other differentiable physics frameworks, with particular attention to scalability and generality. We then conduct ablation studies to evaluate the impact of the techniques presented in Sections~\ref{sec3-2} and~\ref{sec3-3}. Lastly, we provide case studies that illustrate the application of the presented approach to inverse problems and learning control.

The automatic differentiation is implemented in PyTorch 1.3~\cite{Paszke2019}. All experiments are run on an Intel(R) Xeon(R) W-2123 CPU @ 3.60GHz.

\subsection{Scalability}
We construct two benchmark scenes and vary them in controlled fashion to evaluate scalability. The scenes are illustrated in Figure~\ref{fig:scale}.

In the first, objects fall from the air, hit the ground, and finally settle. To test scalability, we increase the number of objects while maintaining the stride between objects. As the number of objects increases, the spatial extent of the scene expands accordingly.

In the second scene, a rigid bunny strikes a deformable cloth. We vary the relative scale of the bunny and the cloth to test the ability of simulation frameworks to handle objects with geometric features at different resolutions.

Few differentiable simulation frameworks are capable of modeling these scenes. \citet{Belbute2018} only simulate 2D scenes with a restricted repertoire of primitives. \citet{Liang2019} simulate cloth but cannot handle rigid-body dynamics. We therefore choose the simulation framework of \citet{Hu2019:ICRA} as our main baseline. This framework uses the the material point method~(MPM), which leverages particles and grids to model solids. We use the state-of-the-art implementation from the high-performance DiffTaichi library~\citep{Hu2019:ICLR}.

The results are shown in Figure~\ref{fig:scale}. The first row reports the runtime and memory consumption of the two frameworks as the number of objects in the scene increases from 20 to 1000. Memory consumption is peak memory usage. Time is the running time for simulating 2 seconds of dynamics. Our runtime and memory consumption increase linearly in the complexity of the scene, while the MPM-based method scales cubically until it runs out of memory at 200 objects and a $640^3$ grid.

The second row reports runtime and memory consumption as the relative sizes of the cloth and the bunny are varied from 1:1 to 10:1. The runtime and memory consumption of our method stay constant. In contrast, as the size of the cloth grows, the MPM-based framework is forced to allocate more and more memory and expend greater and greater computation.

These experiments indicate that scalability is a significant advantage of our method. Since we do not need to quantize space, the extent of the scene or the relative sizes of objects do not dominate our runtime and memory consumption. Since our method dynamically detects and handles collisions locally, as needed, rather than setting up a global optimization problem, the runtime scales linearly (rather than quadratically) with scene complexity.

\subsection{Ablation Studies}

We proposed localized collision handling in Section~\ref{sec3-2} and fast differentiation in Section~\ref{sec3-3} to make our method scalable to large scenes with many degrees of freedom. We conduct dedicated ablation studies to assess the contribution of these techniques.

\begin{figure}
\centering
\begin{tabular}{@{}c@{\hspace{1mm}}c@{}}
    \includegraphics[width=.5\linewidth]{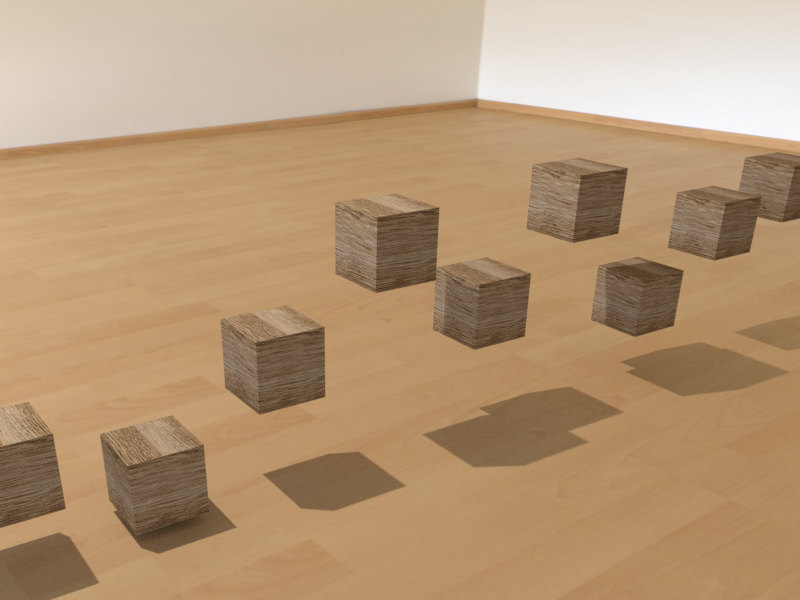} &
    \includegraphics[width=.5\linewidth]{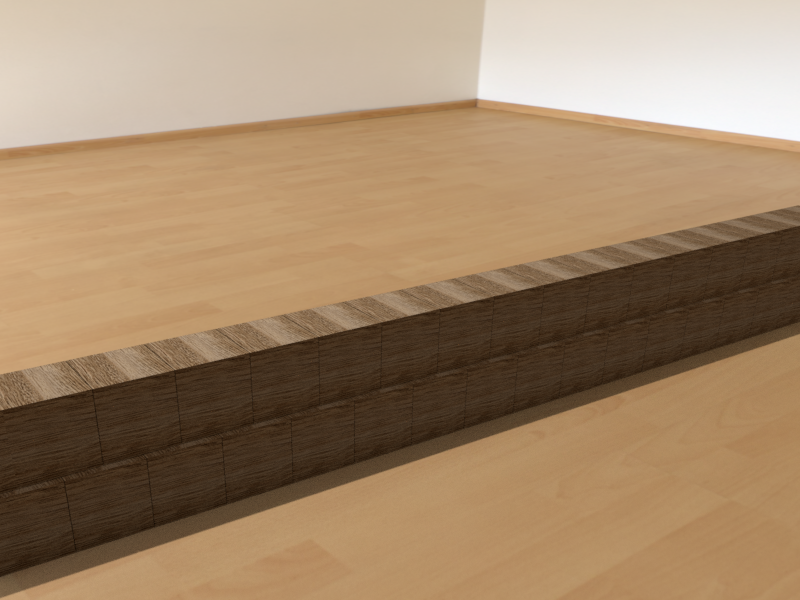} \\
    \small (a) Localized collisions & \small  (b) Correlated collisions
\end{tabular}
\vspace{-4mm}
\caption{Ablation studies. (a) Comparing global and local collision handling. Cubes are released from the air and fall to the ground. (b) Evaluating the contribution of the fast differentiation scheme. Cubes are densely stacked, forming a big connected component. All collisions need to be solved simultaneously because motion of one cube can affect all others.}
\vspace{-3mm}
\label{fig:abeg}
\end{figure}

\mypara{Localized collision handling.}
We compare with LCP-based rigid-body differentiable simulation developed by~\citet{Belbute2018}. This baseline also uses
implicit differentiation to compute derivatives of optimization solvers~\cite{AmosKolter2017}.
The environment is shown in Figure~\ref{fig:abeg}(a). N (= 100, 200, 300) cubes are released above the ground plane. They fall down and hit the ground.
We use the same environment for both methods,
although the LCP-based framework only simulates in 2D, thus having only half the degrees of freedom compared to our 3D simulation (3 versus 6 per object, accounting for translation and rotation).
We disabled our fast differentiation method in this experiment in order to neutralize its effect and conduct a controlled comparison between global and local collision handling.

The implementation of \citet{Belbute2018} uses four threads while our method only uses one.
Results are reported in Table~\ref{tab:ab-sparsity}. Our sparse collision handling method runs up to 5 times faster than the LCP-based approach, and the performance gap widens as the complexity of the scene increases.

\begin{table}[!htb]
\centering
\ra{1.1}
\resizebox{1\linewidth}{!}{
	\small
	\begin{tabular}{@{}l@{\hspace{6mm}}r@{\hspace{4mm}}r@{\hspace{4mm}}r@{\hspace{4mm}}r@{}}
		\toprule
		\# of cubes & 100 & 200  &  300  \\
		\toprule
		LCP    &   0.73s $\pm$ 0.017s &  2.87s $\pm$ 0.103s &  8.42s $\pm$ 0.190s  \\
		Ours    & \textbf{0.56s $\pm$ 0.009s} &\textbf{1.11s $\pm$ 0.012s} &  \textbf{1.65s $\pm$ 0.025s} \\
		\bottomrule
	\end{tabular}
}
\vspace{-4mm}
\caption{Runtime of backpropagation (in seconds per simulation step) with LCP-based collision handling~\citep{Belbute2018} and our approach. Simulation of the scene shown in Figure~\ref{fig:abeg}(a). The LCP-based framework simulates in 2D and uses 4 threads, while our implementation simulates in 3D and uses 1 thread. Our approach is faster, and the performance gap widens with the complexity of the scene.}
\label{tab:ab-sparsity}
\end{table}

\mypara{Fast differentiation.} 
We now evaluate the contribution of the acceleration scheme described in Section~\ref{sec3-3}.
The environment is shown in Figure~\ref{fig:abeg}(b).
N (= 100, 200, 300) cubes are stacked in two layers.
During collision handling, all contacts between cubes form one connected component.
All constraints need to be solved in one big optimization problem.
Thus the linear system in Equation~\ref{eq:system} is large.

\begin{table}[!htb]
\centering
\ra{1.1}
\resizebox{1\linewidth}{!}{
	\small
	\begin{tabular}{@{}l@{\hspace{6mm}}r@{\hspace{4mm}}r@{\hspace{4mm}}r@{\hspace{4mm}}r@{}}
		\toprule
		\# of cubes & 100 & 200 & 300    \\
		\toprule
	W/o FD	   & 1.43s $\pm$ 0.015s & 7.76s $\pm$ 0.302s &   21.88s $\pm$ 0.125s    \\
	Ours	   & \textbf{0.41s $\pm$ 0.004s} & \textbf{0.86s $\pm$ 0.008s} &  \textbf{1.30s $\pm$ 0.008s}  \\
	Speedup  & 3.49x  & 9.02x & 16.83x \\
		\bottomrule
	\end{tabular}
}
\vspace{-4mm}
\caption{Runtime of backpropagation (in seconds per simulation step) with and without our fast differentiation scheme. Simulation of the scene shown in Figure~\ref{fig:abeg}(b).
N ( = 100, 200, 300) cubes are stacked in two layers. 
The impact of the acceleration scheme increases with the complexity of the scene.}
\label{tab:ab-qr}
\end{table}

We evaluate the runtime of backpropagation in this scene with and without the presented acceleration scheme. The ablation condition is referred to as `W/o FD', where FD refers to fast differentiation. In this condition, the derivative of Equation~\ref{eq:reform} is calculated by solving Equation~\ref{eq:system} directly.
The results are reported in Table~\ref{tab:ab-qr}.
The fast differentiation accelerates backpropagation by up to 16x in this scene, and the impact of the technique increases with the complexity of the scene.

\subsection{Two-way coupling}
\begin{figure}
\centering
\begin{tabular}{@{}c@{\hspace{1mm}}c@{}}
    \includegraphics[width=.5\linewidth]{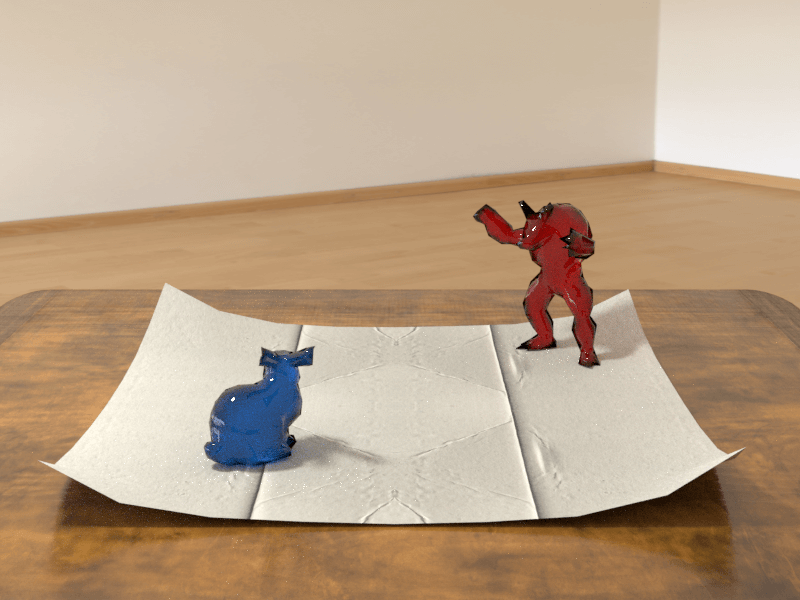} &
    \includegraphics[width=.5\linewidth]{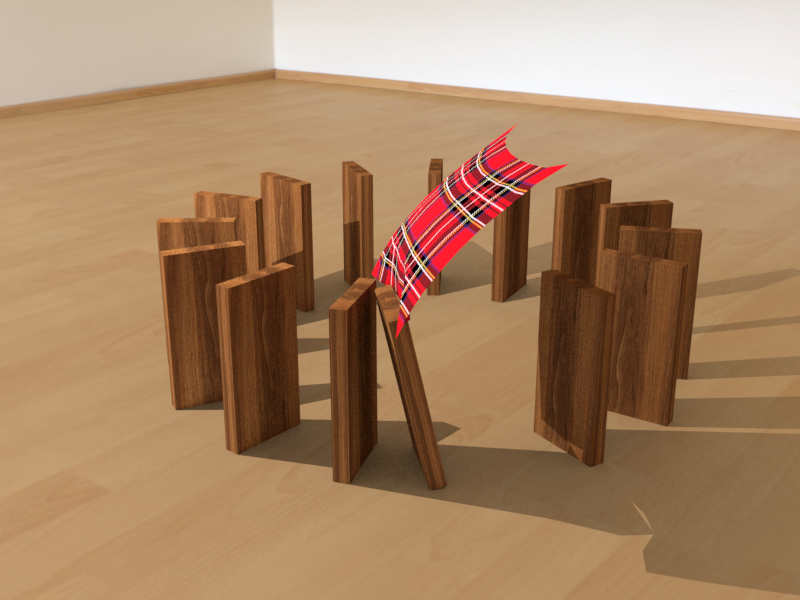} \\
    \includegraphics[width=.5\linewidth]{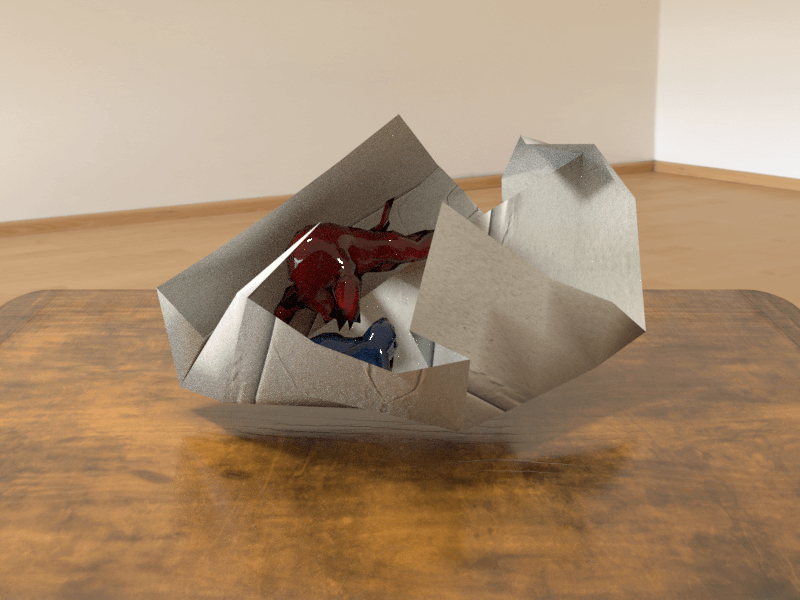} &
    \includegraphics[width=.5\linewidth]{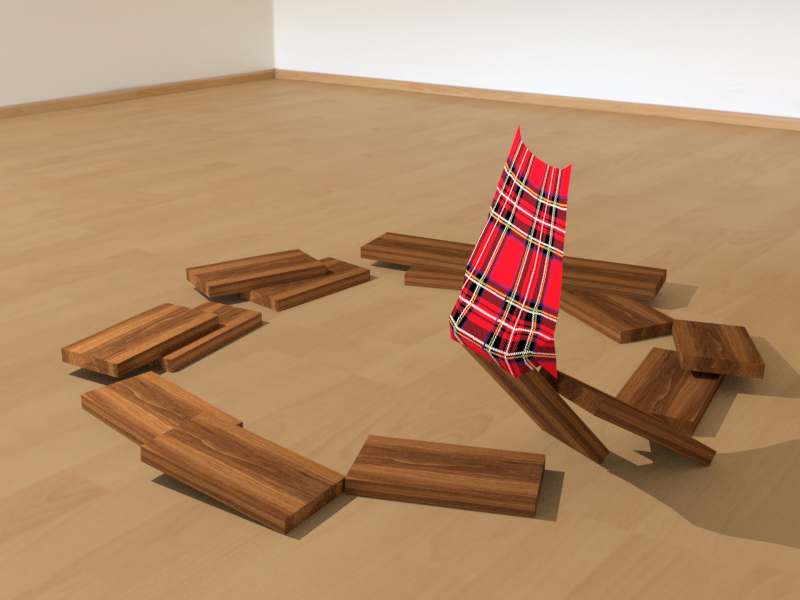} \\
    \small (a) Bunny and armadillo & \small  (b) Dominoes
\end{tabular}
\caption{Two-way coupling of rigid bodies and cloth. Top: initial state. Bottom: towards the end of the simulation. (a) Two figurines are lifted by a piece of cloth. (b) Cloth strikes domino and is later struck back. No prior differentiable simulation framework is capable of modeling these scenes properly.}
\vspace{-3mm}
\label{fig:qualitative}
\end{figure}

One of the advantages of this presented framework is that it can handle both rigid bodies and deformable objects such as cloth. This is illustrated in the two scenes shown in Figure~\ref{fig:qualitative}. In Figure~\ref{fig:qualitative}(a), a Stanford bunny and an armadillo stand on a piece of cloth. The cloth is lifted by its corners, envelopes the figurines, and lifts them up. In Figure~\ref{fig:qualitative}(b), a piece of cloth strikes a domino, which begins a chain reaction that propagates until the last domino strikes the cloth from behind. The scenes are also shown in the supplementary video.

\begin{figure}
\centering
\begin{tabular}{@{}c@{\hspace{1mm}}c@{}}
    \includegraphics[width=.5\linewidth]{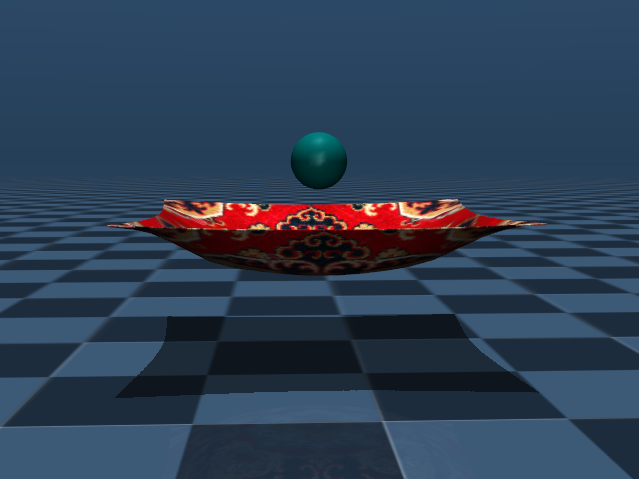} &
    \includegraphics[width=.5\linewidth]{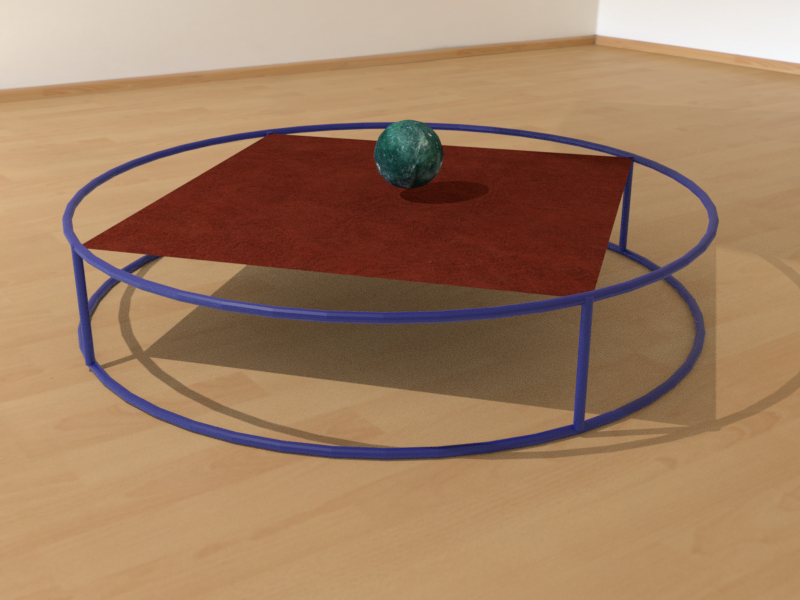} \\
    \includegraphics[width=.5\linewidth]{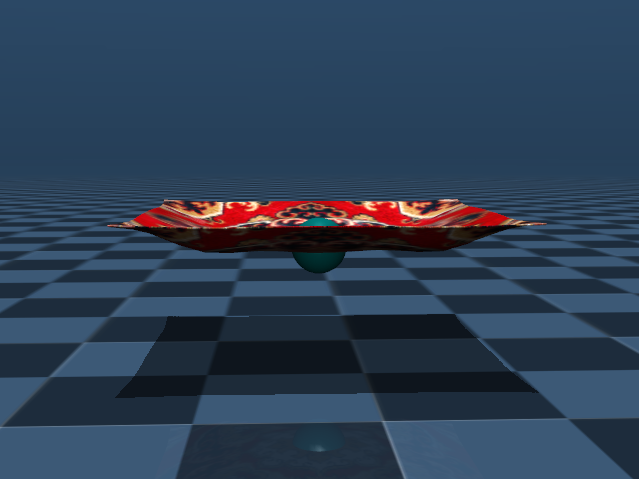} &
    \includegraphics[width=.5\linewidth]{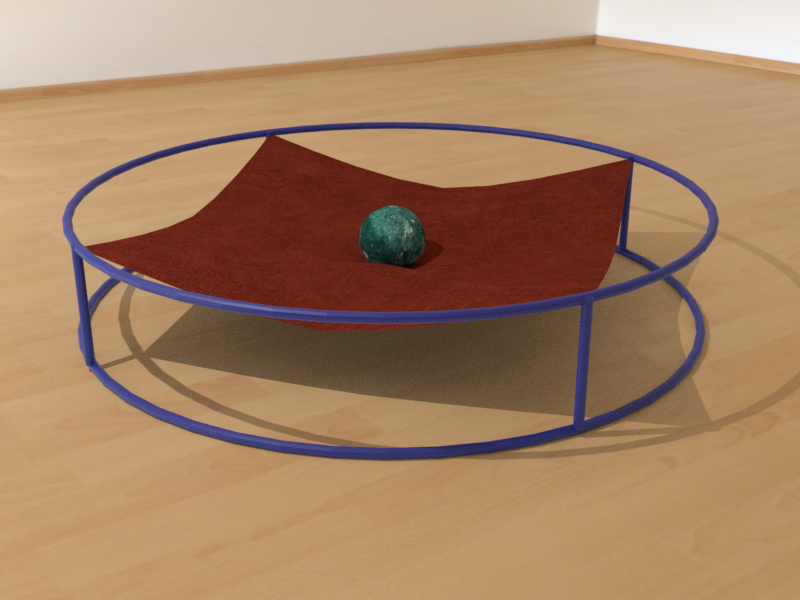} \\
    \small (a) MuJoCo & \small  (b) Ours
\end{tabular}
\vspace{-3mm}
\caption{Comparison with MuJoCo. Simulation of a ball interacting with a trampoline. Top: initial state. Bottom: later in the simulation. (a) Simulation in MuJoCo. (b) Simulation by our method. MuJoCo models cloth as a 2D grid of capsules and ellipsoid geoms in addition to spheres. The ball penetrates the trampoline when the grid is sparse.}
\vspace{-3mm}
\label{fig:mujoco}
\end{figure}

No prior differentiable simulation framework is capable of modeling these scenes properly. \citet{Liang2019} can simulate the cloth, but its motion does not affect the figurines. The framework of \citet{Hu2019:ICRA,Hu2019:ICLR} does not enforce rigidity and suffers from interpenetration. In contrast, our approach simulates the correct dynamics and the two-way coupling between them.

We further compare with MuJoCo, a popular physics engine~\cite{todorov2012mujoco}. MuJoCo models cloth as a 2D grid of capsule and ellipsoid geoms in addition to spheres. This representation fails to correctly handle collisions near the holes in a grid. In Figure~\ref{fig:mujoco}, we simulate a rigid body falling onto a trampoline and bouncing back. When simulated in MuJoCo, the ball penetrates the trampoline. Our method simulates the interaction correctly.

\subsection{Applications}

\begin{figure}
\centering
\begin{tabular}{@{}c@{\hspace{1mm}}c@{}}
    \includegraphics[width=0.5\linewidth]{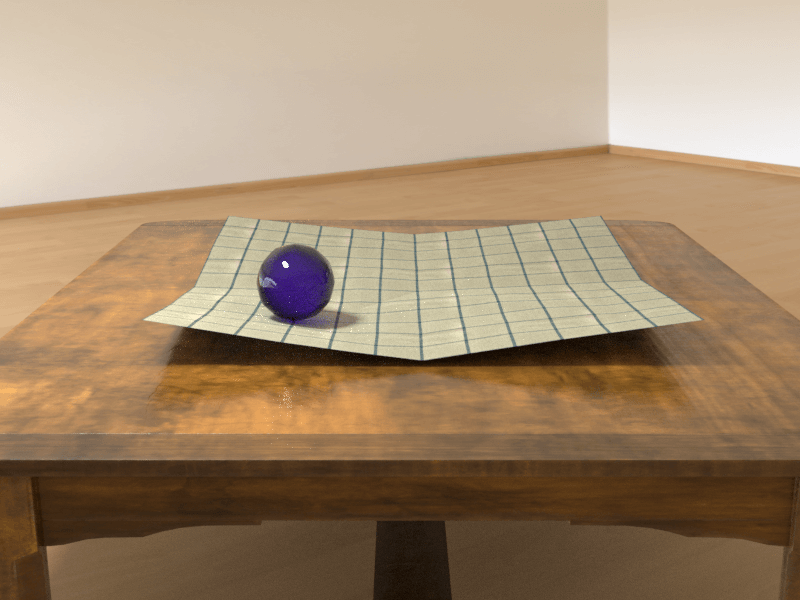} &
    \includegraphics[width=0.5\linewidth]{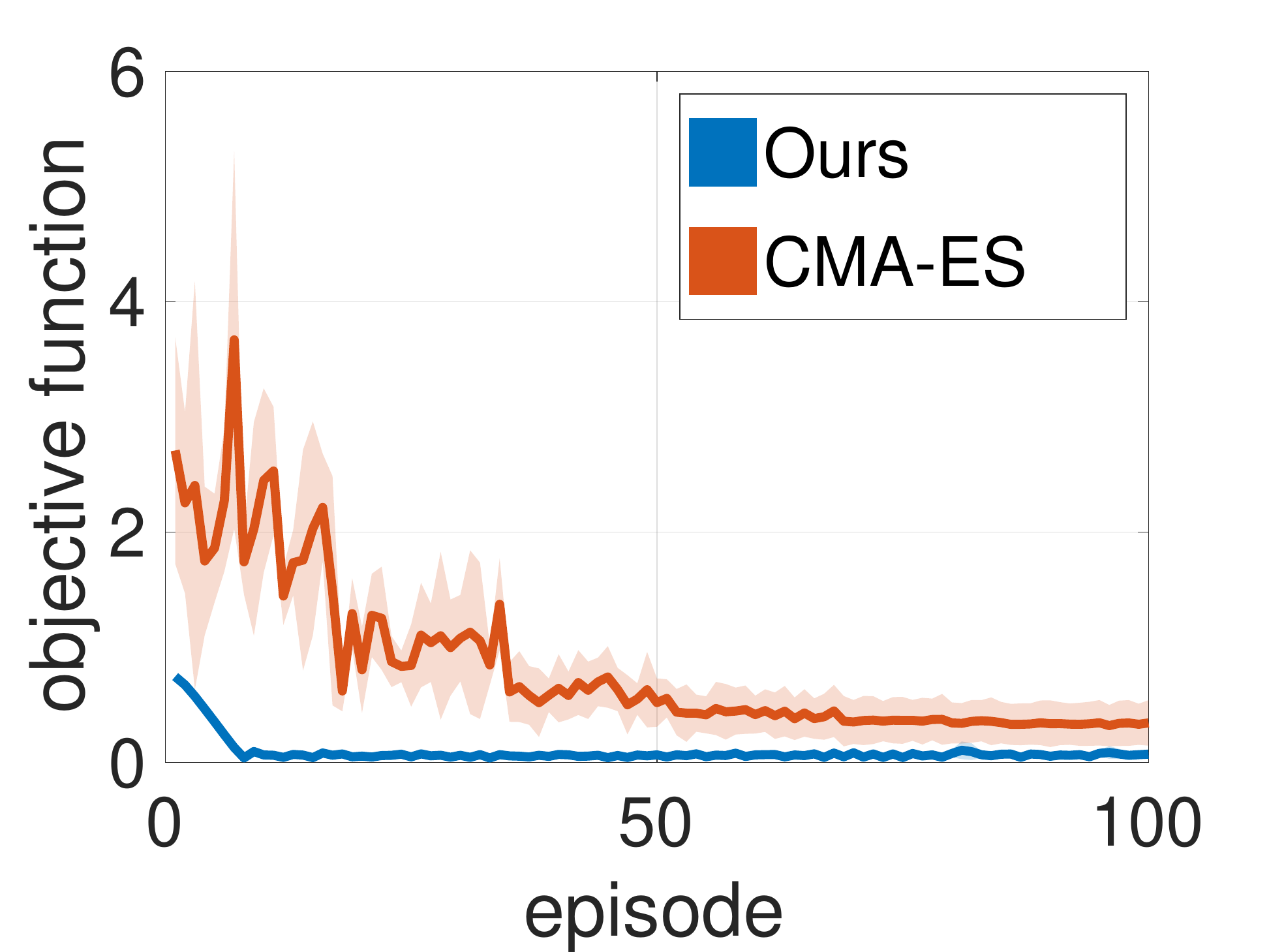} \\
 
    \small (a) Marble on a sheet & \small (b) Objective function 
\end{tabular}
\vspace{-4mm}
\caption{Inverse problem. (a) The goal is to bring the marble to the desired position with minimal force. (b) Our framework enables gradient-based optimization for this problem and quickly converges to a lower objective value than derivative-free optimization (CMA-ES). We perform 5 runs with different random seeds; shaded areas represent one standard deviation. 
}
\label{fig:inverse}
\vspace{-4mm}
\end{figure}

\mypara{Inverse problem.}
Differentiable simulation naturally lends itself to gradient-based optimization for inverse problems. 
A case study is shown in Figure~\ref{fig:inverse}, in which a marble is supported by a soft sheet. The sheet is fixed at the four corners.
In each time step, an external force is applied to the marble. The goal is to find a sequence of forces that drives the marble to a target position in 2 seconds, while minimizing the total amount of applied force.
The vertical component of the external force is set to 0 so that the marble has to interact with the cloth before reaching the target.
We compare with a derivative-free optimization algorithm, CMA-ES~\cite{HansenTutorial}. Our framework enables gradient-based optimization in this setting and converges in 4 iterations, reaching a lower objective value than what CMA-ES achieves after two orders of magnitude more iterations.

\begin{figure}[H]
\centering
\begin{tabular}{@{}c@{\hspace{1mm}}c@{}}
    \includegraphics[width=.5\linewidth]{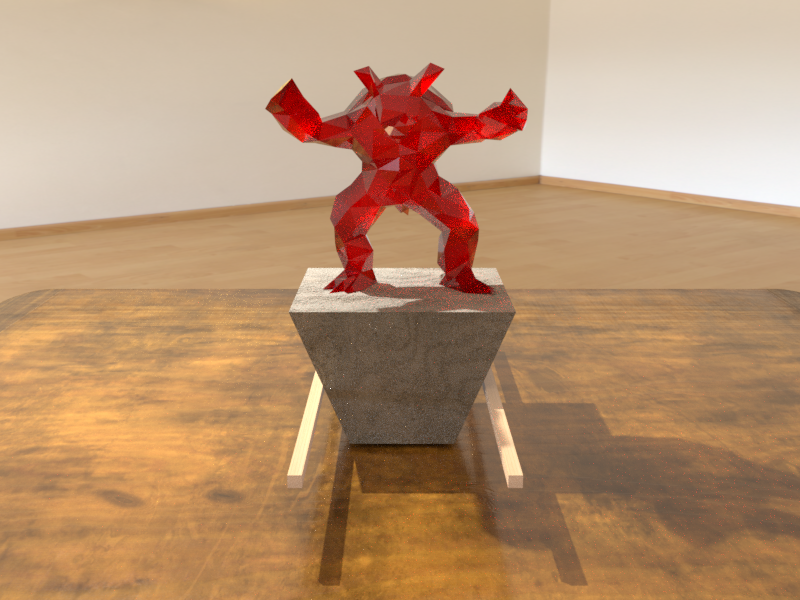} &
    \includegraphics[width=.5\linewidth]{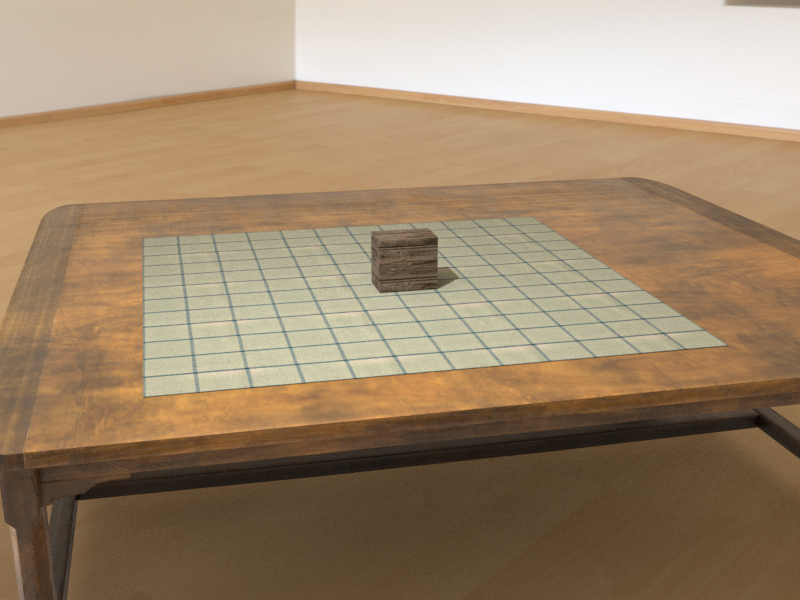} \\
    \includegraphics[width=.5\linewidth]{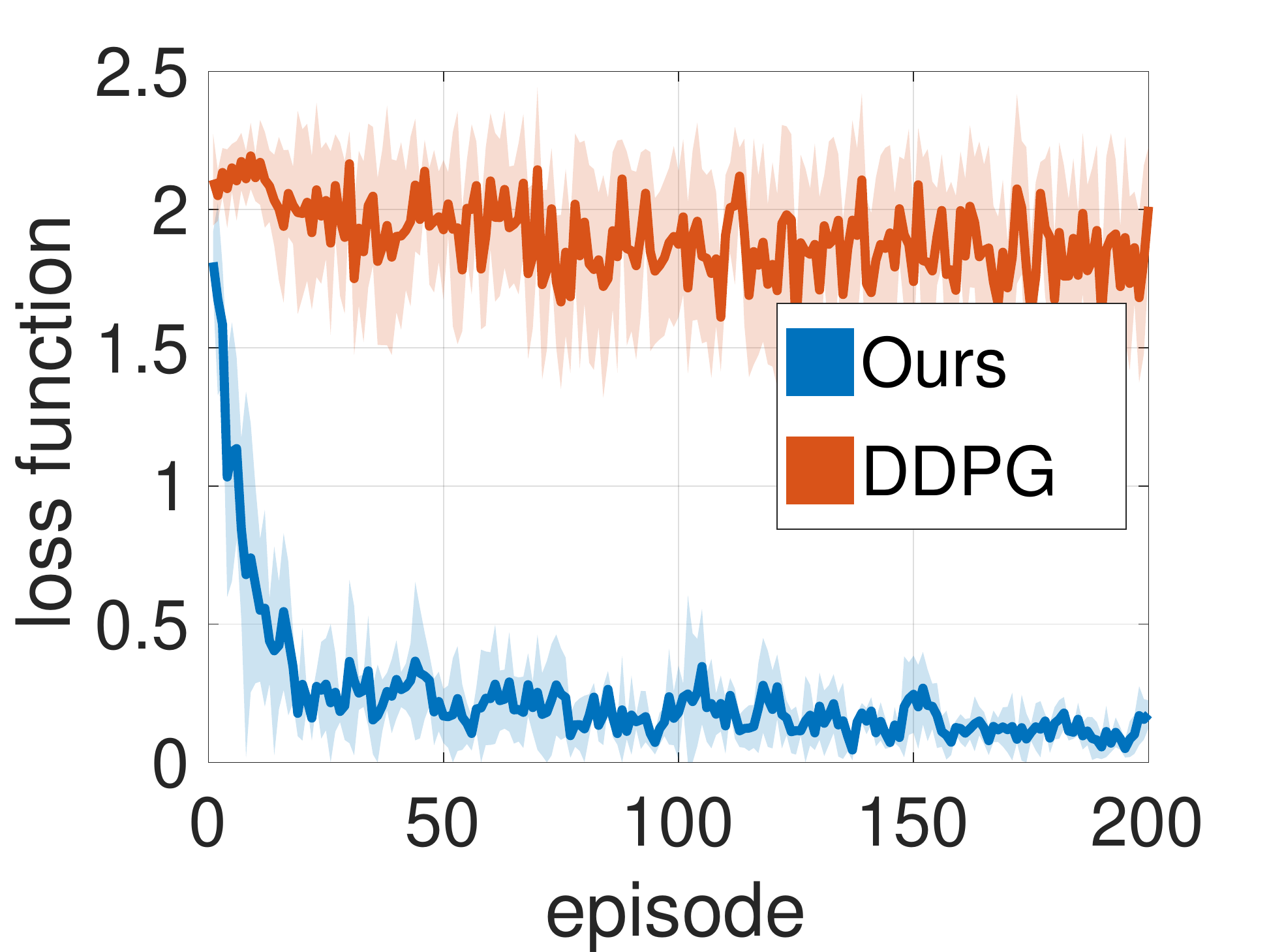} &
    \includegraphics[width=.5\linewidth]{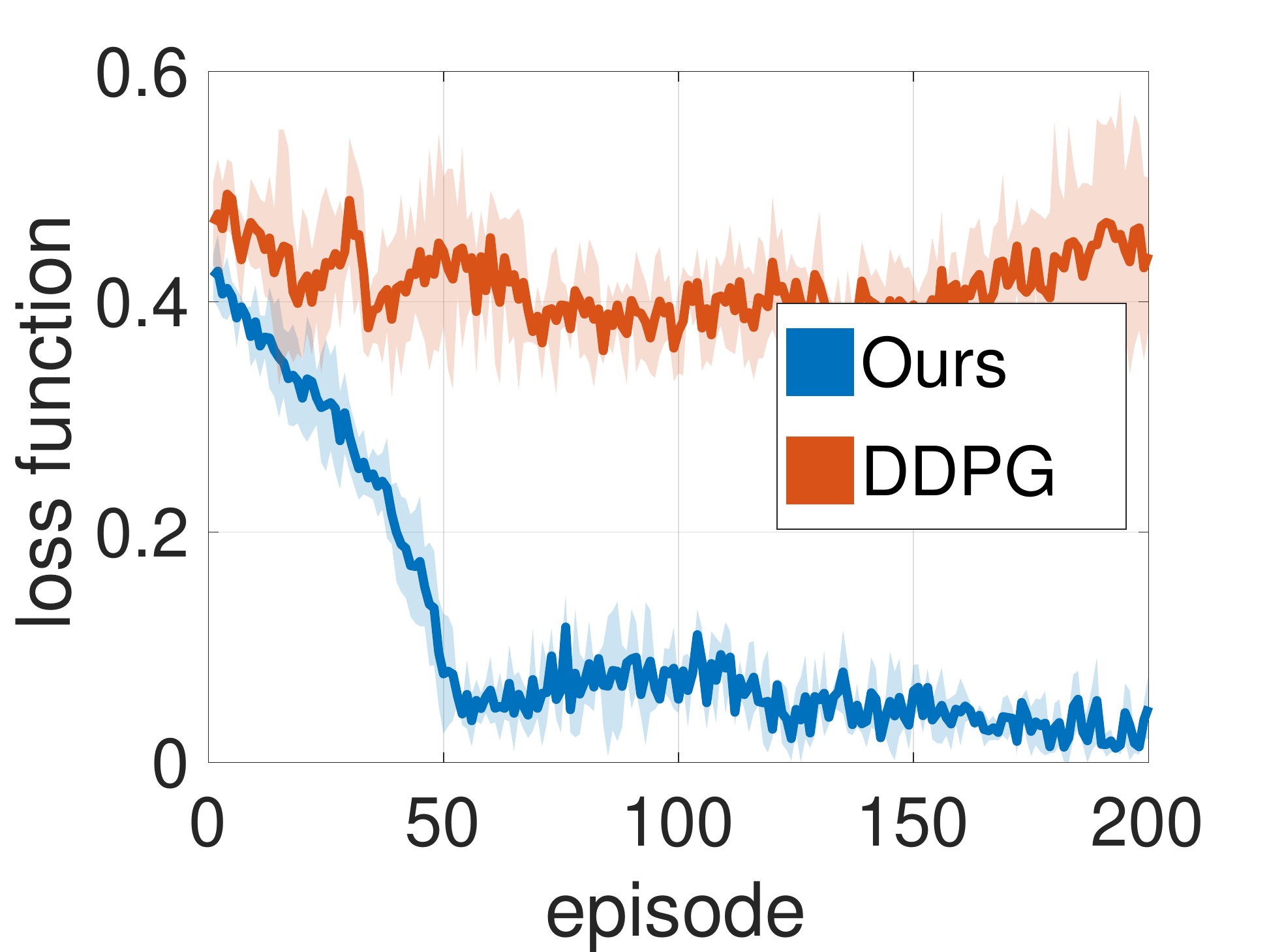} \\
    \small (a) Sticks & \small (b) Cloth
\end{tabular}
\vspace{-3mm}
\caption{Learning control. A neural network must learn to control (a) a pair of sticks or (b) a piece of cloth. The trained controller must be able to bring the manipulated object to any desired location specified at test time. Our method enables gradient-based training of the controller. A model-free reinforcement learning baseline (DDPG) fails to learn the task on a comparable time scale. We perform 5 runs with different random seeds; shaded areas represent one standard deviation.
}
\label{fig:manipulate}
\vspace{-3mm}
\end{figure}

\mypara{Learning control.}
Our second set of case studies uses the differentiable simulator to backpropagate gradients in a neural network that must learn to manipulate objects to achieve desired outcomes. The scenarios are shown in Figure~\ref{fig:manipulate}. The neural network controls a pair of sticks and a piece of cloth, respectively, which must be used to manipulate an object. The goal is to bring the object to the desired position within 1 second. The optimization objective is the $L_2$ distance to the target. The neural network is an MLP with 50 nodes in the first layer and 200 nodes in the second, with ReLU activations.

For reference, we report the performance of DDPG, a model-free reinforcement learning algorithm~\cite{Lillicrap2015}. In each episode, we fix the initial position of the manipulator and the object while the target position is randomized. For both methods, the input to the network is a concatenated vector of relative distance to the target, speed, and remaining time. The DDPG reward is the negative $L_2$ distance to the target.

Our method updates the network once at the end of each episode, while DDPG receives a reward signal and updates the network weights in each time step. As shown in the loss curves in Figure~\ref{fig:manipulate}, our method quickly converges to a good control policy, while DDPG fails to learn the task on a comparable time scale. This example illustrates the power of gradient-based optimization with our differentiable physics framework.

\begin{figure}
\centering
\begin{tabular}{@{}c@{\hspace{1mm}}c@{}}
    \includegraphics[width=0.5\linewidth]{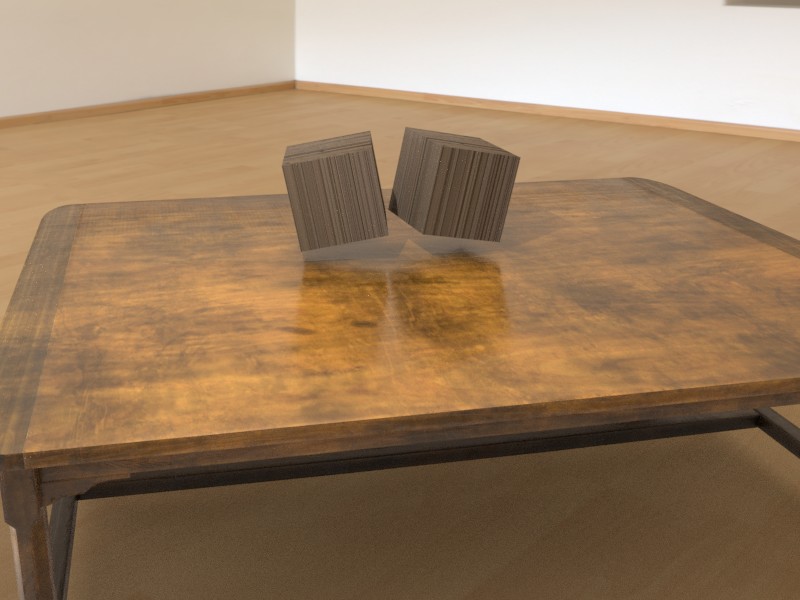} &
    \includegraphics[width=0.5\linewidth]{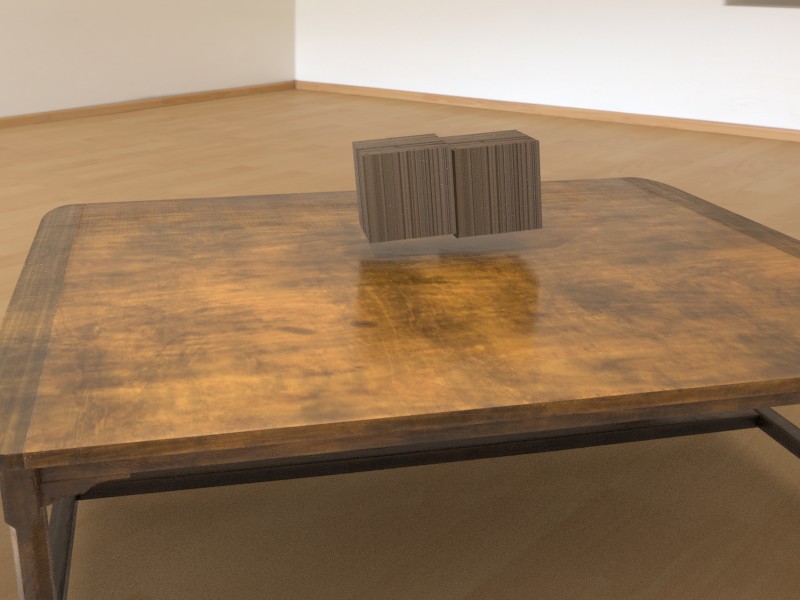} \\
    \small (a) Initial guess & \small (b) Target observation 
\end{tabular}
\vspace{-3mm}
\caption{Parameter estimation. Cubes start with opposite velocities and collide with each other. Our goal is to estimate the mass of the left cube such that the total momentum after the collision has the desired direction and magnitude. (a)~is the initial estimate: both cubes have the same mass and the total momentum after collision is 0. (b)~is the simulation with the mass produced by our estimation procedure: the momentum points to the right and has a magnitude of 3, as specified in the target observation.
}
\vspace{-3mm}
\label{fig:estimate}
\end{figure}

\begin{figure}
\centering
\begin{tabular}{@{}c@{\hspace{1mm}}c@{}}
    \includegraphics[width=0.5\linewidth]{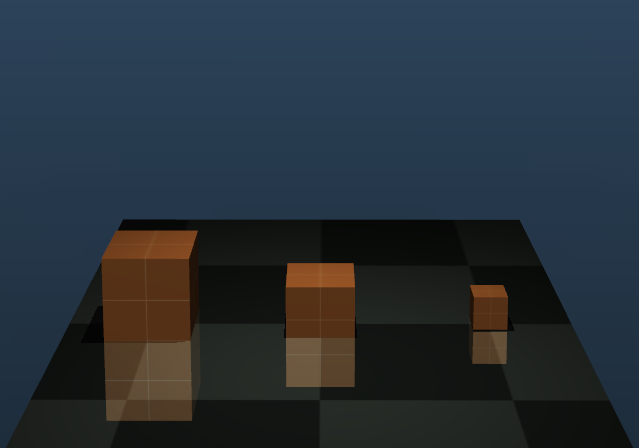} &
    \includegraphics[width=0.5\linewidth]{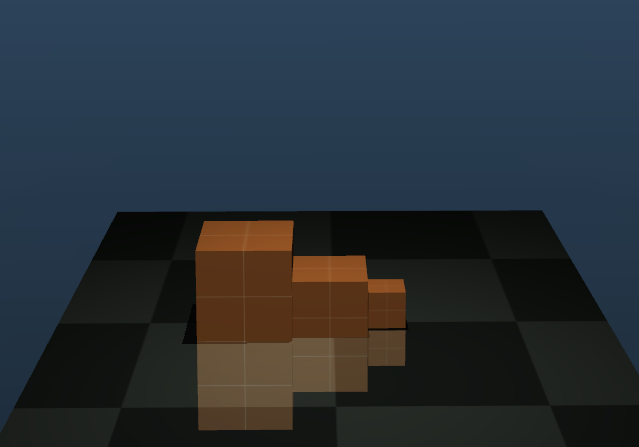} \\
    \small (a) Initial state & \small (b) Target state 
\end{tabular}
\vspace{-3mm}
\caption{
Interoperability across simulators. Three cubes are placed on smooth ground. The goal is to make the cubes stick together while minimizing the applied forces. The loss is computed in MuJoCo but the gradient is evaluated in DiffSim. (a) is the initial state and (b) is the successful simulation result after the optimization.
}
\label{fig:mismatch}
\vspace{-3mm}
\end{figure}

\balance

\mypara{Parameter estimation.}
Given an observation of motion, our method can estimate unknown physical parameters for objects in the scene.

In Figure~\ref{fig:estimate}, two cubes start with opposite velocities $\pm\vv$ and collide with each other. Given a target momentum observation $\pp=m_1 \vv_1'+m_2 \vv_2'$, where $\vv_{1,2}'$ are velocities after collision and $m_{1,2}$ are masses, we aim to estimate the mass $m_1$. The target momentum is set to $\pp=(3,0,0)$. We initialize with $m_1=m_2=1$ and $\pp=(0,0,0)$ (Figure~\ref{fig:estimate}(a)). After 90 gradient steps, our method arrives at the estimate $m_1=5.4$ and achieves the desired momentum (Figure~\ref{fig:estimate}(b)).

\mypara{Interoperability across simulators.}
Our differentiable simulation framework is interoperable with other physics simulators. We have evaluated this interoperability with MuJoCo, a popular non-differentiable simulator~\cite{todorov2012mujoco}. The experiment is illustrated in Figure~\ref{fig:mismatch}. We place three cubes on smooth ground. The goal is to apply forces to the cubes in order to make them stick together. The objective function is the distance of each cube to its target positions plus the $L_2$ norm of the force. We compute the loss in MuJoCo but evaluate the gradient in DiffSim, our differentiable simulator. Figure~\ref{fig:mismatch}(a) shows the initial state. Figure~\ref{fig:mismatch}(b) is the result of the simulation after 10 gradient steps. The goal of making the cubes stick together was accomplished. This experiment shows that physical states and control signals are interoperable between our differentiable framework and non-differentiable simulators.

\section{Conclusion}
\label{sec:conclusion}

We developed a scalable differentiable simulation method that has linear complexity with respect to the number of objects and collisions, and constant complexity with respect to spatial resolution. We use meshes as our core representation, enabling support for objects of arbitrary shape at any scale. A unified local collision handling scheme enables fast coupling between objects with different materials. An acceleration scheme speeds up backpropagation by implicit differentiation. Experiments show that the presented techniques speed up backpropagation by large multiplicative factors, and their impact increases with the complexity of simulated scenes.

We demonstrated the application of the presented method in a number of case studies, including end-to-end gradient-based training of neural network controllers. Training with differentiable simulation was shown to be significantly more effective than gradient-free and model-free baselines.

One avenue for future work is to extend the presented framework to other types of objects, materials, and contact models. The framework is sufficiently general to support deformable solids and articulated bodies, as long as they have mesh-based representations with repulsive contact forces. We plan to extend our implementation to incorporate such object classes.

\mypara{Acknowledgement:} This research is support in part by Intel Corporation, National Science Foundation, and Elizabeth Stevinson Iribe Endowed Chair Professorship.

{\small
	\bibliographystyle{icml2020}
	\bibliography{paper}

\begin{thebibliography}{30}
\providecommand{\natexlab}[1]{#1}
\providecommand{\url}[1]{\texttt{#1}}
\expandafter\ifx\csname urlstyle\endcsname\relax
  \providecommand{\doi}[1]{doi: #1}\else
  \providecommand{\doi}{doi: \begingroup \urlstyle{rm}\Url}\fi

\bibitem[Amos \& Kolter(2017)Amos and Kolter]{AmosKolter2017}
Amos, B. and Kolter, J.~Z.
\newblock {OptNet}: Differentiable optimization as a layer in neural networks.
\newblock In \emph{ICML}, 2017.

\bibitem[Battaglia et~al.(2016)Battaglia, Pascanu, Lai, Rezende, and
  Kavukcuoglu]{battaglia2016interaction}
Battaglia, P.~W., Pascanu, R., Lai, M., Rezende, D., and Kavukcuoglu, K.
\newblock Interaction networks for learning about objects, relations and
  physics.
\newblock In \emph{Neural Information Processing Systems}, 2016.

\bibitem[Botsch et~al.(2010)Botsch, Kobbelt, Pauly, Alliez, and
  L{\'e}vy]{Botsch2010}
Botsch, M., Kobbelt, L., Pauly, M., Alliez, P., and L{\'e}vy, B.
\newblock \emph{Polygon mesh processing}.
\newblock AK Peters/CRC Press, 2010.

\bibitem[Bridson et~al.(2002)Bridson, Fedkiw, and Anderson]{Bridson2002}
Bridson, R., Fedkiw, R., and Anderson, J.
\newblock Robust treatment of collisions, contact and friction for cloth
  animation.
\newblock \emph{{ACM} Trans. Graph.}, 2002.

\bibitem[Carpentier \& Mansard(2018)Carpentier and
  Mansard]{CarpentierMansard2018}
Carpentier, J. and Mansard, N.
\newblock Analytical derivatives of rigid body dynamics algorithms.
\newblock In \emph{Robotics: Science and Systems (RSS)}, 2018.

\bibitem[Chang et~al.(2017)Chang, Ullman, Torralba, and
  Tenenbaum]{chang2017compositional}
Chang, M.~B., Ullman, T., Torralba, A., and Tenenbaum, J.~B.
\newblock A compositional object-based approach to learning physical dynamics.
\newblock In \emph{ICLR}, 2017.

\bibitem[de~Avila~Belbute{-}Peres et~al.(2018)de~Avila~Belbute{-}Peres, Smith,
  Allen, Tenenbaum, and Kolter]{Belbute2018}
de~Avila~Belbute{-}Peres, F., Smith, K.~A., Allen, K.~R., Tenenbaum, J., and
  Kolter, J.~Z.
\newblock End-to-end differentiable physics for learning and control.
\newblock In \emph{Neural Information Processing Systems}, 2018.

\bibitem[Degrave et~al.(2019)Degrave, Hermans, Dambre, and
  wyffels]{Degrave2019}
Degrave, J., Hermans, M., Dambre, J., and wyffels, F.
\newblock A differentiable physics engine for deep learning in robotics.
\newblock \emph{Frontiers in Neurorobotics}, 13, 2019.

\bibitem[Hansen(2016)]{HansenTutorial}
Hansen, N.
\newblock The {CMA} evolution strategy: {A} tutorial.
\newblock \emph{arXiv:1604.00772}, 2016.

\bibitem[Harmon et~al.(2008)Harmon, Vouga, Tamstorf, and Grinspun]{Harmon2008}
Harmon, D., Vouga, E., Tamstorf, R., and Grinspun, E.
\newblock Robust treatment of simultaneous collisions.
\newblock \emph{{ACM} Trans. Graph.}, 2008.

\bibitem[Holl et~al.(2020)Holl, Koltun, and Thuerey]{Holl2020}
Holl, P., Koltun, V., and Thuerey, N.
\newblock Learning to control {PDEs} with differentiable physics.
\newblock In \emph{ICLR}, 2020.

\bibitem[Hu et~al.(2019)Hu, Liu, Spielberg, Tenenbaum, Freeman, Wu, Rus, and
  Matusik]{Hu2019:ICRA}
Hu, Y., Liu, J., Spielberg, A., Tenenbaum, J.~B., Freeman, W.~T., Wu, J., Rus,
  D., and Matusik, W.
\newblock {ChainQueen}: {A} real-time differentiable physical simulator for
  soft robotics.
\newblock In \emph{ICRA}, 2019.

\bibitem[Hu et~al.(2020)Hu, Anderson, Li, Sun, Carr, Ragan{-}Kelley, and
  Durand]{Hu2019:ICLR}
Hu, Y., Anderson, L., Li, T., Sun, Q., Carr, N., Ragan{-}Kelley, J., and
  Durand, F.
\newblock {DiffTaichi}: Differentiable programming for physical simulation.
\newblock In \emph{ICLR}, 2020.

\bibitem[Ingraham et~al.(2019)Ingraham, Riesselman, Sander, and
  Marks]{ingraham2019learning}
Ingraham, J., Riesselman, A., Sander, C., and Marks, D.
\newblock Learning protein structure with a differentiable simulator.
\newblock In \emph{ICLR}, 2019.

\bibitem[Li et~al.(2019{\natexlab{a}})Li, Wu, Tedrake, Tenenbaum, and
  Torralba]{Li2019}
Li, Y., Wu, J., Tedrake, R., Tenenbaum, J.~B., and Torralba, A.
\newblock Learning particle dynamics for manipulating rigid bodies, deformable
  objects, and fluids.
\newblock In \emph{ICLR}, 2019{\natexlab{a}}.

\bibitem[Li et~al.(2019{\natexlab{b}})Li, Wu, Zhu, Tenenbaum, Torralba, and
  Tedrake]{Li2019:ICRA}
Li, Y., Wu, J., Zhu, J.-Y., Tenenbaum, J.~B., Torralba, A., and Tedrake, R.
\newblock Propagation networks for model-based control under partial
  observation.
\newblock In \emph{ICRA}, 2019{\natexlab{b}}.

\bibitem[Liang et~al.(2019)Liang, Lin, and Koltun]{Liang2019}
Liang, J., Lin, M.~C., and Koltun, V.
\newblock Differentiable cloth simulation for inverse problems.
\newblock In \emph{Neural Information Processing Systems}, 2019.

\bibitem[Lillicrap et~al.(2016)Lillicrap, Hunt, Pritzel, Heess, Erez, Tassa,
  Silver, and Wierstra]{Lillicrap2015}
Lillicrap, T.~P., Hunt, J.~J., Pritzel, A., Heess, N., Erez, T., Tassa, Y.,
  Silver, D., and Wierstra, D.
\newblock Continuous control with deep reinforcement learning.
\newblock In \emph{ICLR}, 2016.

\bibitem[Macklin et~al.(2014)Macklin, M{\"{u}}ller, Chentanez, and
  Kim]{Macklin2014}
Macklin, M., M{\"{u}}ller, M., Chentanez, N., and Kim, T.
\newblock Unified particle physics for real-time applications.
\newblock \emph{{ACM} Trans. Graph.}, 2014.

\bibitem[Millard et~al.(2020)Millard, Heiden, Agrawal, and
  Sukhatme]{Millard2020}
Millard, D., Heiden, E., Agrawal, S., and Sukhatme, G.~S.
\newblock Automatic differentiation and continuous sensitivity analysis of
  rigid body dynamics.
\newblock \emph{arXiv:2001.08539}, 2020.

\bibitem[Mrowca et~al.(2018)Mrowca, Zhuang, Wang, Haber, Li, Tenenbaum, and
  Yamins]{Mrowca2018}
Mrowca, D., Zhuang, C., Wang, E., Haber, N., Li, F., Tenenbaum, J., and Yamins,
  D.~L.
\newblock Flexible neural representation for physics prediction.
\newblock In \emph{Neural Information Processing Systems}, 2018.

\bibitem[Narain et~al.(2012)Narain, Samii, and O'Brien]{Narain2012}
Narain, R., Samii, A., and O'Brien, J.~F.
\newblock Adaptive anisotropic remeshing for cloth simulation.
\newblock \emph{{ACM} Trans. Graph.}, 2012.

\bibitem[Paszke et~al.(2019)Paszke, Gross, Massa, Lerer, Bradbury, Chanan,
  Killeen, Lin, Gimelshein, Antiga, et~al.]{Paszke2019}
Paszke, A., Gross, S., Massa, F., Lerer, A., Bradbury, J., Chanan, G., Killeen,
  T., Lin, Z., Gimelshein, N., Antiga, L., et~al.
\newblock {PyTorch}: An imperative style, high-performance deep learning
  library.
\newblock In \emph{Neural Information Processing Systems}, 2019.

\bibitem[Sanchez-Gonzalez et~al.(2018)Sanchez-Gonzalez, Heess, Springenberg,
  Merel, Riedmiller, Hadsell, and Battaglia]{Sanchez-Gonzalez2018}
Sanchez-Gonzalez, A., Heess, N., Springenberg, J.~T., Merel, J., Riedmiller,
  M., Hadsell, R., and Battaglia, P.
\newblock Graph networks as learnable physics engines for inference and
  control.
\newblock In \emph{ICML}, 2018.

\bibitem[Schenck \& Fox(2018)Schenck and Fox]{Schenck2018}
Schenck, C. and Fox, D.
\newblock {SPNets}: Differentiable fluid dynamics for deep neural networks.
\newblock In \emph{Conference on Robot Learning (CoRL)}, 2018.

\bibitem[Spielberg et~al.(2019)Spielberg, Zhao, Hu, Du, Matusik, and
  Rus]{Spielberg2019}
Spielberg, A., Zhao, A., Hu, Y., Du, T., Matusik, W., and Rus, D.
\newblock Learning-in-the-loop optimization: End-to-end control and co-design
  of soft robots through learned deep latent representations.
\newblock In \emph{Neural Information Processing Systems}, 2019.

\bibitem[Todorov et~al.(2012)Todorov, Erez, and Tassa]{todorov2012mujoco}
Todorov, E., Erez, T., and Tassa, Y.
\newblock {MuJoCo}: A physics engine for model-based control.
\newblock In \emph{IEEE/RSJ International Conference on Intelligent Robots and
  Systems}, 2012.

\bibitem[Toussaint et~al.(2018)Toussaint, Allen, Smith, and
  Tenenbaum]{Toussaint2018differentiable}
Toussaint, M., Allen, K., Smith, K., and Tenenbaum, J.
\newblock Differentiable physics and stable modes for tool-use and manipulation
  planning.
\newblock In \emph{Robotics: Science and Systems (RSS)}, 2018.

\bibitem[Ummenhofer et~al.(2020)Ummenhofer, Prantl, Th{\"u}rey, and
  Koltun]{Ummenhofer2020}
Ummenhofer, B., Prantl, L., Th{\"u}rey, N., and Koltun, V.
\newblock Lagrangian fluid simulation with continuous convolutions.
\newblock In \emph{ICLR}, 2020.

\bibitem[Witkin \& Baraff(1997)Witkin and Baraff]{Witkin1997}
Witkin, A. and Baraff, D.
\newblock Physically based modeling: Principles and practice.
\newblock {SIGGRAPH} Course Notes, 1997.

\end{thebibliography}
}

\clearpage
\begin{appendices}
\label{sec:appendix}
\section{Mass Matrix}\label{sup:mass}
For a rigid body $i$ with rotation $\rr=(\phi,\theta,\psi)\trans$ and translation $\tt=(t_x,t_y,t_z)\trans$, the generalized coordinates are $\qq=[\rr\trans,\tt\trans]\trans\in\mathbb{R}^6$. Its kinetic energy can be computed by $\EE=\frac{1}{2}\dot{\qq}\hat{\MM}\dot{\qq}$, where $\hat{\MM}$ is a diagonal blocked mass matrix composed of angular and linear inertia,

\begin{equation}
\hat{\MM}=\left[
\begin{array}{cc}
\mathcal{I}_{a} & \bm{0}\\
\bm{0} & \mathcal{I}_{l}
\end{array}\right].
\label{eq:system-app}
\end{equation}

The linear inertia $\mathcal{I}_{l}$ is simply $m\II_{3\times 3}$, where $m$ is the total mass of the rigid body. 

When the rigid body's distribution is approximated by a set of particles, an angular inertial is given by,
\begin{equation}
    \mathcal{I}'= \sum_im_i(\pp_i\trans\pp_i\II_{3\times3}-\pp_i\pp_i\trans),
\end{equation}
where $m_i$ is the mass of the particle $i$; $\pp_i$ is the vector from the center of mass to this particle. Note that this angular inertia corresponds to the axis-angle velocity $\omega$ in the \textit{world frame}. The angular momentum is $\mathcal{I}'\omega$. However, We cannot directly use this formula since we have to represent the angular velocity in terms of velocities of Euler angles.

We used the RPY convention for Euler angle representation: given the Euler angle $\rr=(\phi,\theta,\psi)\trans$, the rigid body will first rotate about the $Z$ axis by $\psi$, then rotate about the new $Y'$ axis by $\theta$, and finally rotate about the new $X''$ axis by $\phi$.

By the above definition, the angular velocity vector $\Omega$ can be represented by Euler angles,
\begin{equation}
    \Omega=\dot{\psi}\ee_{Z}+\dot{\theta}\ee_{Y'}+\dot{\phi}\ee_{X''}
\end{equation}

We convert the angular velocity in local frame back to the world frame so that it matches with the angular inertia in the world frame:
\begin{align}
    \Omega&=(\cos{\theta}\cos{\psi}\dot{\phi}-\sin{\psi}\dot{\theta})\ee_X \notag\\
    &+(\cos{\theta}\cos{\psi}\dot{\phi}+\cos{\psi}\dot{\theta})\ee_Y. \notag\\
    &+(-\sin{\theta}\dot{\phi}+\dot{\psi})\ee_Z
\end{align}

The matrix form is given as:
\begin{equation}
\left[
\begin{array}{c}
\omega_x\\
\omega_y\\
\omega_z
\end{array}\right]
=
\left[
\begin{array}{ccc}
\cos{\theta}\cos{\psi} & -\sin{\psi} & 0\\
\cos{\theta}\sin{\psi} & \cos{\psi} & 0\\
-\sin{\theta} & 0 & 1
\end{array}\right]
\left[
\begin{array}{c}
\dot{\phi}\\
\dot{\theta}\\
\dot{\psi}
\end{array}\right].
\end{equation}

We denote this transformation as $\omega=\TT\dot{\rr}$, so the angular momentum is reformed as $\mathcal{I}'\omega=\mathcal{I}'\TT\dot{\rr}$. Therefore, in the Euler angle representation, the angular inertia becomes
\begin{equation}
    \mathcal{I}_{a}=\TT\trans\mathcal{I}'\TT.
\end{equation}

The new mass matrix $\hat{\MM}$ for the generalized coordinates is,

\begin{equation}
\hat{\MM}=\left[
\begin{array}{cc}
\TT\trans\mathcal{I}'\TT & \bm{0}\\
\bm{0} & \II_{3\times3}
\end{array}\right].
\label{eq:system-app2}
\end{equation}

\section{Represent a Vertex using Generalized Coordinates}\label{sup:trans}
For a vertex $p$ attached to the rigid body, its coordinate in the body frame is $\pp_0$. The origin of the body frame is set to be the center of mass (COM). $\pp_0=(p_x,p_y,p_z)\trans$ is the relative displacement of $p$ w.r.t. the COM in the first frame of simulation. The coordinates of the vertex in the world frame are then given by,
\begin{equation}\label{eq:f}
    \pp=\bm{f}(\qq)=[\rr]\pp_0+\tt,
\end{equation}
where $[\rr]$ is a rotation matrix represented by the Euler angle $\rr=(\phi,\theta,\psi)\trans$. The corresponding rotation matrix $[\rr]=\RR_{3\times3}$ is computed by 
$$\RR_{11}=\cos{\theta}\cos{\psi}$$
$$\RR_{12}=-\cos{\phi}\sin{\psi}+\sin{\phi}\sin{\theta}\cos{\psi}$$
$$\RR_{13}=\sin{\phi}\sin{\psi}+\cos{\phi}\sin{\theta}\cos{\psi}$$
$$\RR_{21}=\cos{\theta}\sin{\psi}$$
$$\RR_{22}= \cos{\phi}\cos{\psi}+\sin{\phi}\sin{\theta}\sin{\psi}$$
$$\RR_{23}= -\sin{\phi}\cos{\psi}+\cos{\phi}\sin{\theta}\sin{\psi}$$
$$\RR_{31}=-\sin{\theta} $$
$$\RR_{32}= \sin{\phi}\cos{\theta}  $$
$$\RR_{33}= \cos{\phi}\cos{\theta} $$

\section{Computation of Derivatives}\label{sup:gradient}
To backpropagate the gradients at vertex $\pp$ to the generalized coordinates $\qq$, we need to compute the derivatives $\partial \bm{f}(\qq)/\partial \qq$. 

The coordinates of a vertex $(x,y,z)\trans=\bm{f}([\phi, \theta, \psi, t_x,t_y,t_z]\trans)$ are given by Equation~\ref{eq:f}. Therefore, the Jacobian matrix for the partial derivatives is:

\begin{equation}
 \nabla\bm{f}=
\left[
\begin{array}{cccccc}
\PD{x}{\phi}    & \PD{x}{\theta}  & \PD{x}{\psi} & 1 & 0 & 0\\
\PD{y}{\phi} & \PD{y}{\theta}    & \PD{y}{\psi}  & 0 & 1 & 0\\
\PD{z}{\phi}  & \PD{z}{\theta} & \PD{z}{\psi}    & 0 & 0 & 1
\end{array}
\right].
\label{eq:grada}
\end{equation}

\begin{figure*}
\centering
\begin{tabular}{@{}c@{\hspace{1mm}}c@{\hspace{1mm}}c@{\hspace{1mm}}c@{}}
    \includegraphics[height = 2.4cm]{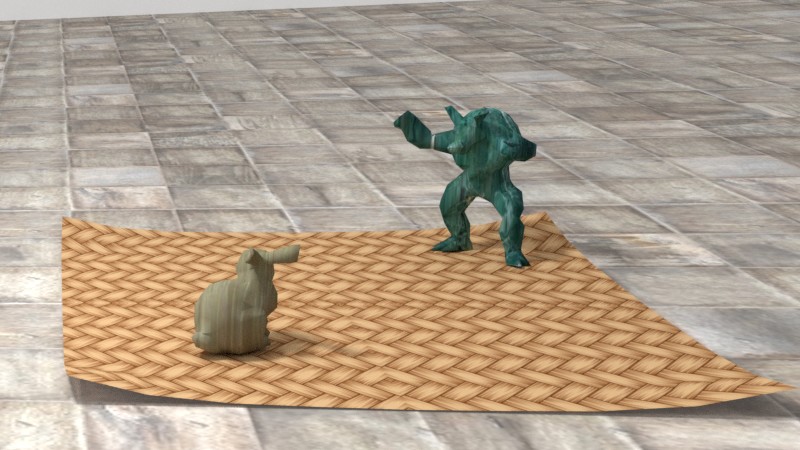} &
    \includegraphics[height = 2.4cm]{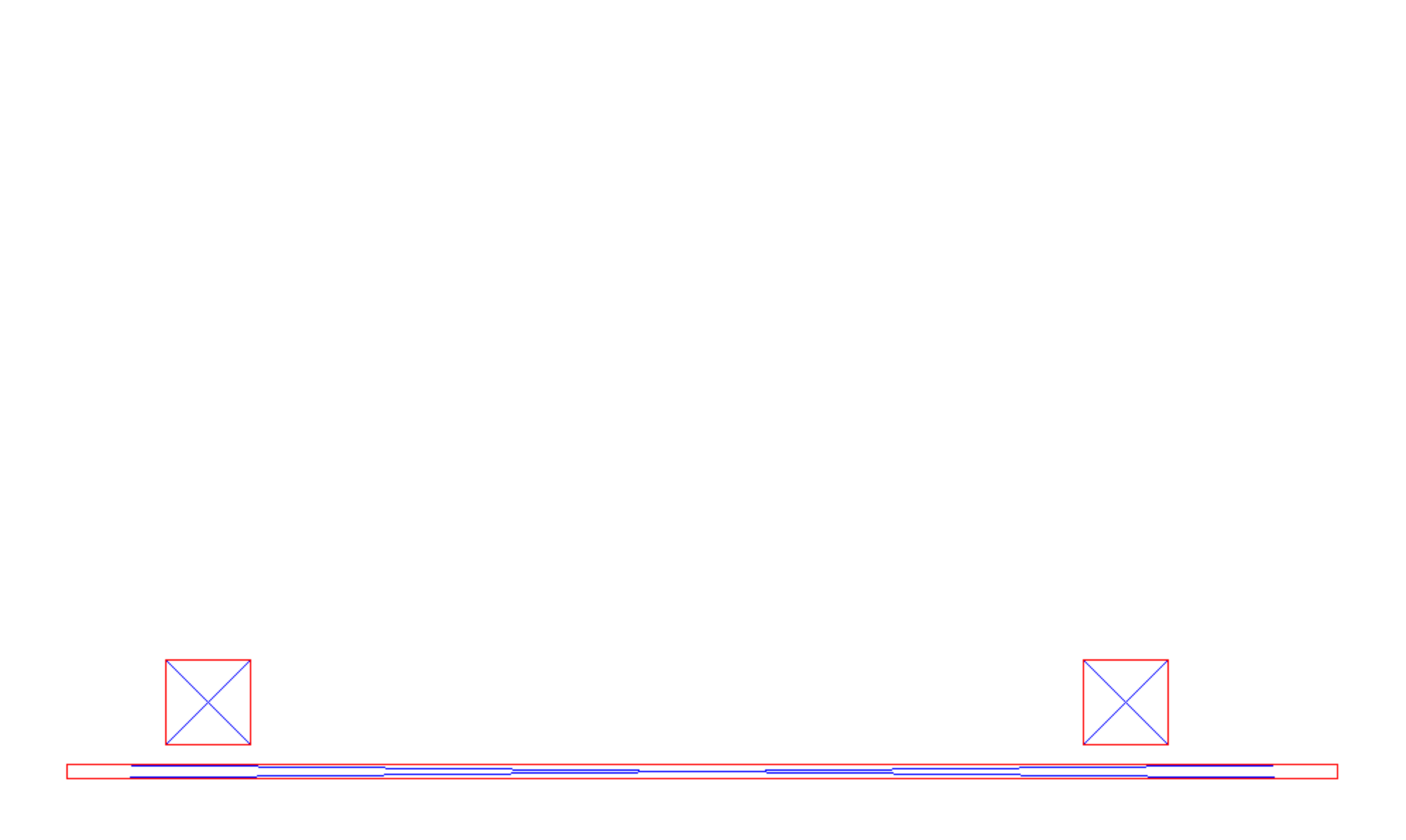} &
    \includegraphics[height = 2.4cm]{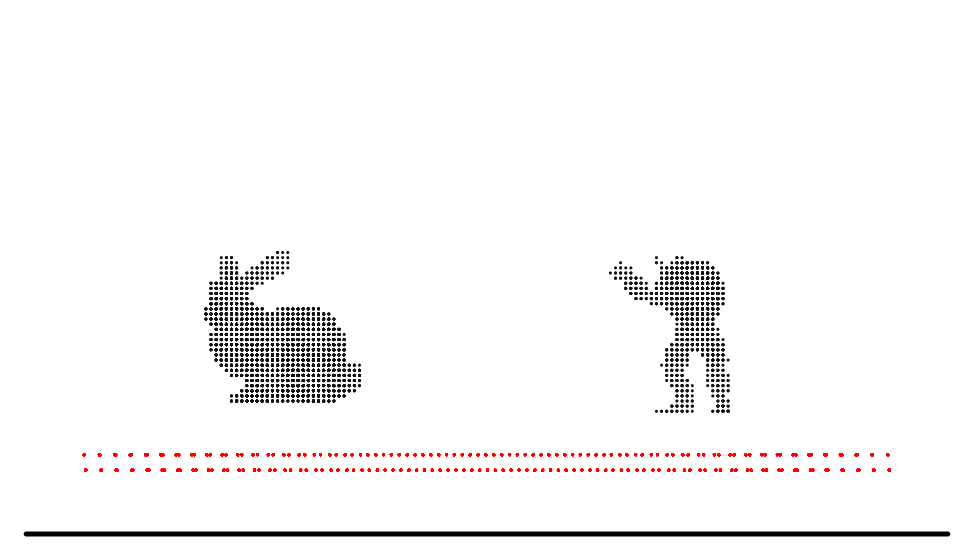} &
    \includegraphics[height = 2.4cm]{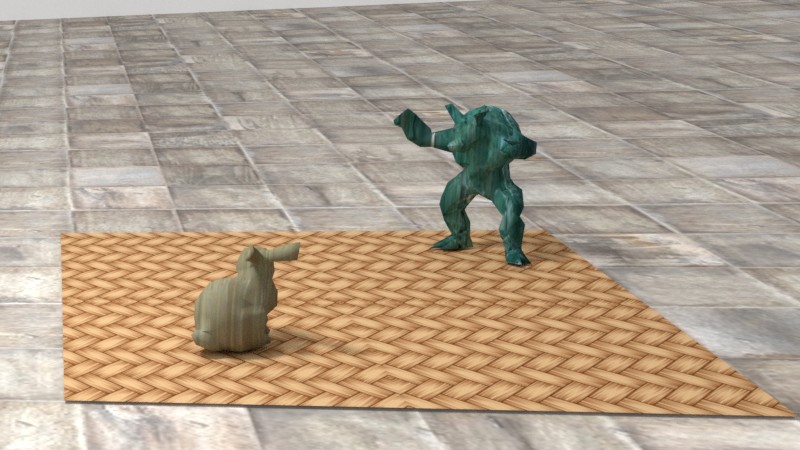} \\
    \includegraphics[height = 2.4cm]{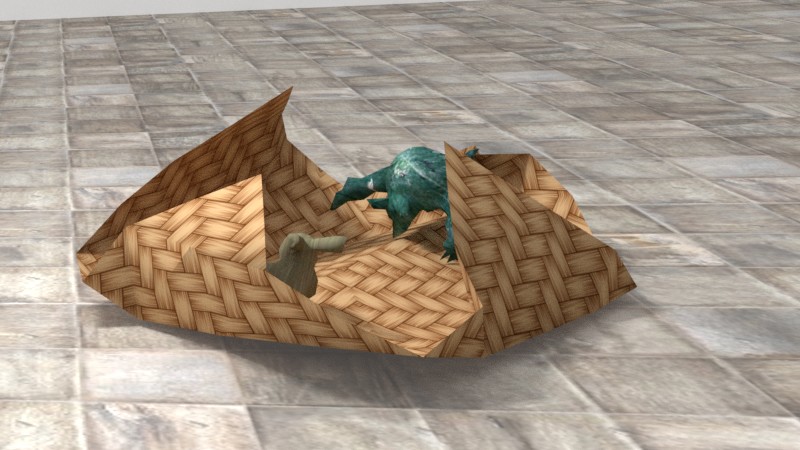} &
    \includegraphics[height = 2.4cm]{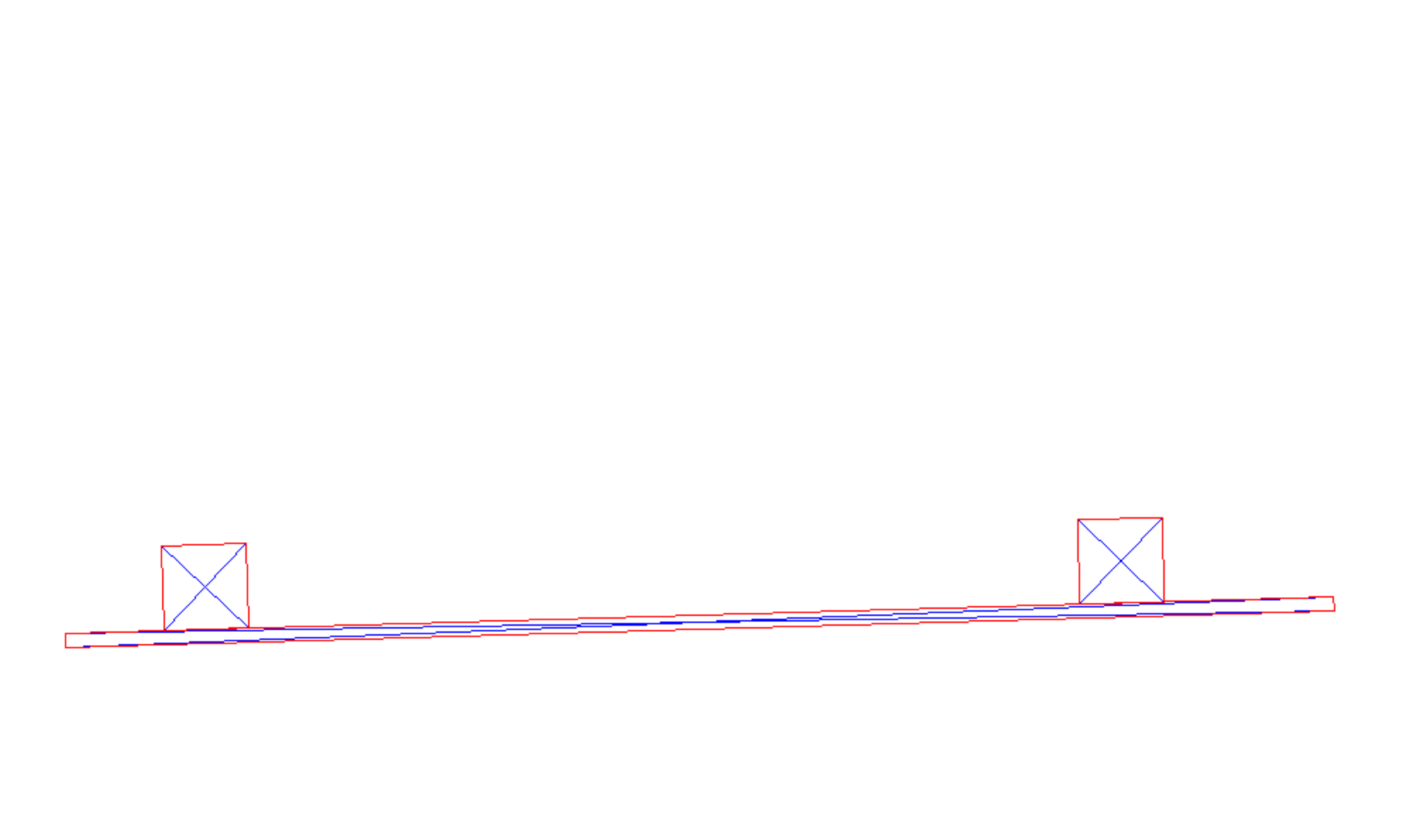} &
    \includegraphics[height = 2.4cm]{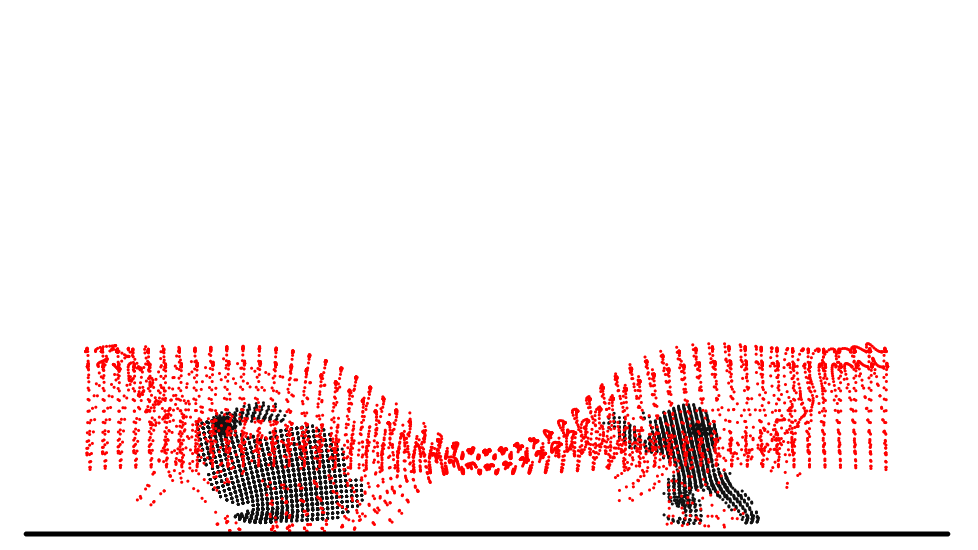} &
    \includegraphics[height = 2.4cm]{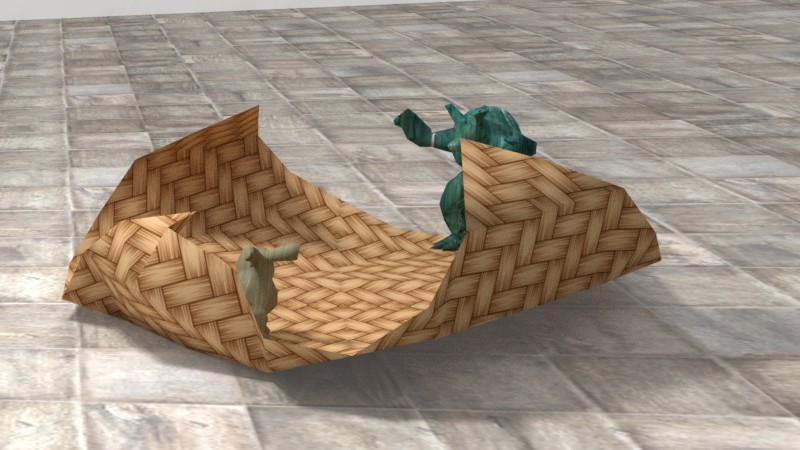} \\
    \includegraphics[height = 2.4cm]{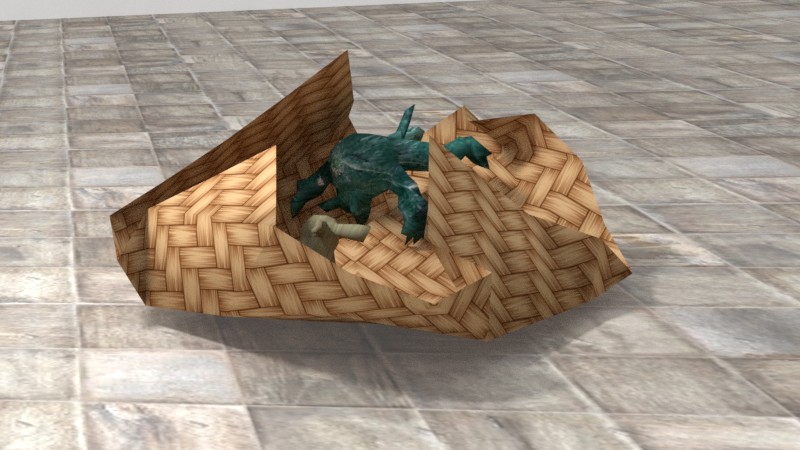} &
    \includegraphics[height = 2.4cm]{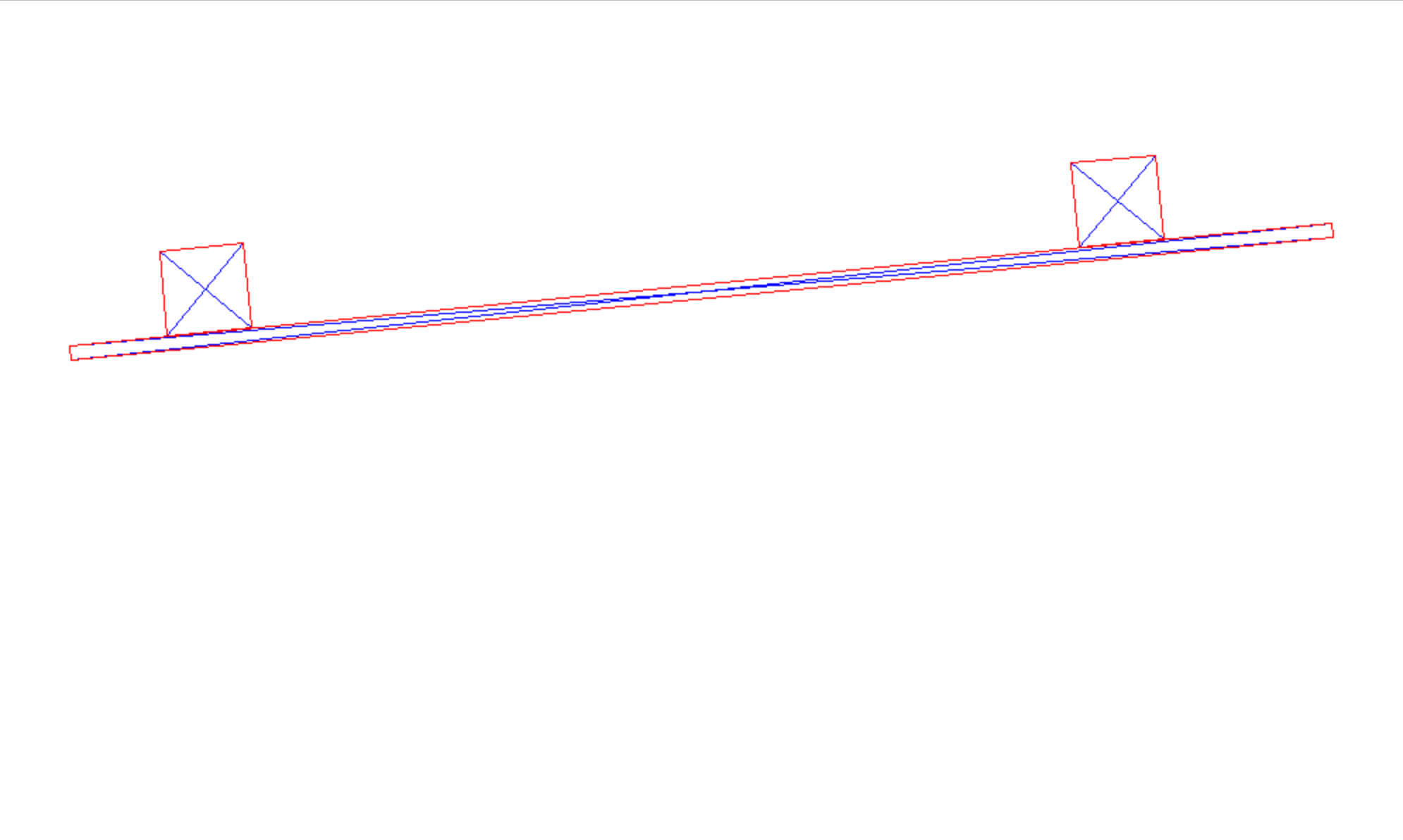} &
    \includegraphics[height = 2.4cm]{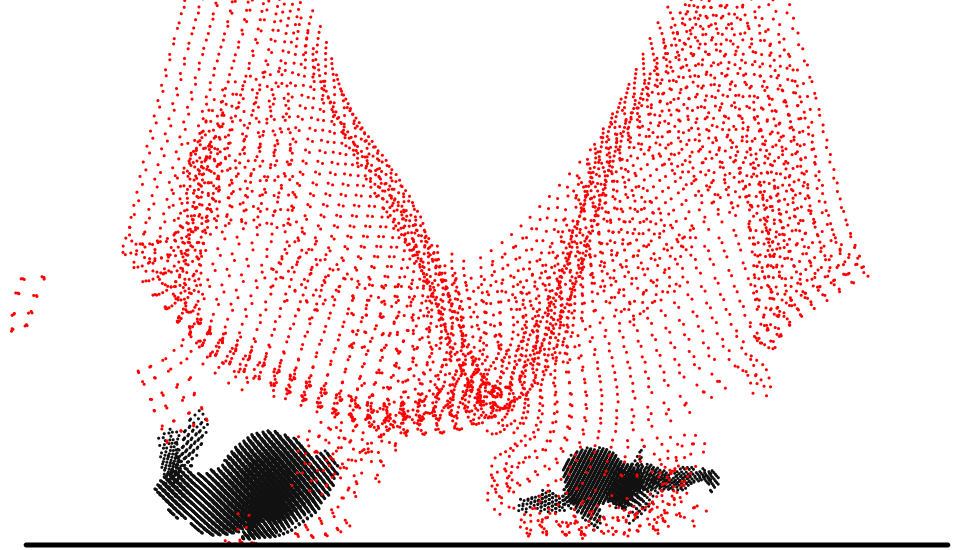} &
    \includegraphics[height = 2.4cm]{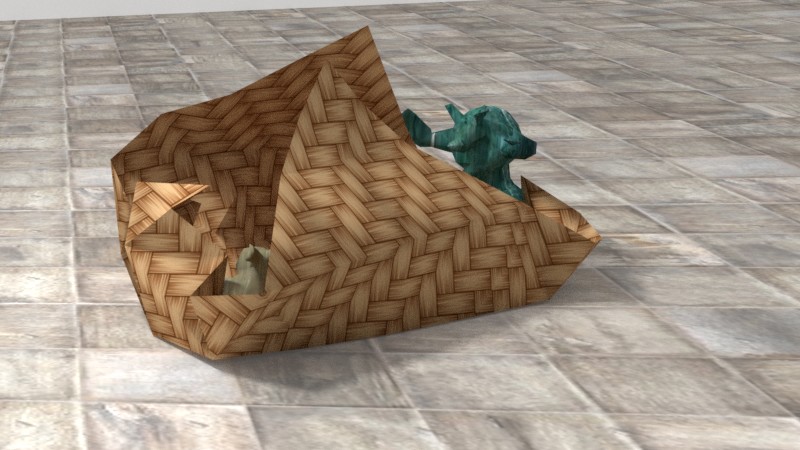} \\
    \small (a) Ours & \small  (b) E2E  & \small (c) ChainQueen  & \small (d) DiffCloth
\end{tabular}
\caption{Qualitative comparison on complex models and interactions. 
In this scene, we simulate two complex rigid bodies, bunny and armadillo on a cloth. (a) is the simulation result of our method, and the comparisons, (b) is E2E~\cite{Belbute2018}; (c) is ChainQueen~\cite{Hu2019:ICRA}; (d) is DiffCloth \cite{Liang2019}. Our method gets the most reasonable results because we can deal with both rigid body and cloth dynamics. Meanwhile, the mesh representation enables us to model different objects precisely.
}
\label{fig:suppcomp}
\end{figure*}

Assuming that the relative displacement of $p$ w.r.t. the COM in the first frame is $\pp_0=(p_x,p_y,p_z)$, the corresponding elements in the Jacobian matrix are:

$\PD{x}{\psi}=-p_x\cos\theta\sin\psi  +p_y(-\cos\phi\cos\psi-\sin\phi\sin\theta\sin\psi)  + p_z (\sin\phi\cos\psi-\cos\phi\sin\theta\sin\psi)$

$\PD{x}{\theta}= -p_x\sin\theta\cos\psi+p_y\sin\phi\cos\theta\cos\psi+p_z \cos\phi\cos\theta\cos\psi$

$\PD{x}{\phi}=p_y \sin\phi\sin\psi+\cos\phi\sin\theta\cos\psi+p_z (\cos\phi\sin\psi-\sin\phi\sin\theta\cos\psi)$

$\PD{y}{\psi}= p_x \cos\theta\cos\psi+ p_y(-\cos\phi\sin\psi+\sin\phi\sin\theta\cos\psi)+p_z (\sin\phi\sin\psi+\cos\phi\sin\theta\cos\psi)$

$\PD{y}{\theta}= -p_x \sin\theta\sin\psi +p_y \sin\phi\cos\theta\sin\psi +p_z \cos\phi\cos\theta\sin\psi$

$\PD{y}{\phi}=p_y (-\sin\phi\cos\psi+\cos\phi\sin\theta\sin\psi) + p_z (-\cos\phi\cos\psi-\sin\phi\sin\theta\sin\psi)$

$\PD{z}{\psi}=0$

$\PD{z}{\theta}=-p_x \cos\theta-p_y\sin\phi\sin\theta-p_z \cos\phi\sin\theta$

$\PD{z}{\phi}=p_y \cos\phi\cos\theta-p_z\sin\phi\cos\theta$

\section{Qualitative Comparisons}\label{sup:comp}
Figure~\ref{fig:suppcomp} shows qualitative comparisons of two complex rigid bodies interacting with a soft cloth. 

In this comparison, our differentiable physics gets the most realistic results, that both the bunny and armadillo are lifted from the table. E2E cannot simulate the detailed motion of the objects because it only supports basic primitives. ChainQueen cannot preserve the \textit{rigid} property of bunny and armadillo because these objects are composed of small particles during the simulation and obey deformable body dynamics. Moreover, the cloth would break up because the MPM method cannot keep connectivity information faithfully during the simulation. Diffcloth is able to model the motion of cloth, but the cloth cannot have an impact on the bunny and armadillo, therefore they can not be lifted.

\end{appendices}

\end{document}